\documentclass[lettersize,journal]{IEEEtran}
\usepackage{epsfig}
\usepackage{amsfonts, amssymb}
\usepackage{amsmath,amssymb,amsfonts}
\usepackage{algorithm}
\usepackage{algpseudocode}
\algrenewcommand\algorithmicrequire{\textbf{Input:}}
\algrenewcommand\algorithmicensure{\textbf{Output:}}
\usepackage{array}
\usepackage[caption=false,font=normalsize,labelfont=sf,textfont=sf]{subfig}
\usepackage{textcomp}
\usepackage{stfloats}
\usepackage{url}
\usepackage{verbatim}
\usepackage{graphicx}
\usepackage{multirow}
\usepackage{makecell}
\usepackage{booktabs}
\usepackage{cite}
\usepackage{times}
\usepackage{bbm}
\usepackage{hyperref}
\usepackage{xcolor}
\hypersetup{
	colorlinks=true,
	linkcolor=[rgb]{0,0,0}, % 深蓝色
	citecolor=[rgb]{0,0,0},
	urlcolor=[rgb]{0.0,0.27,0.68}
}

\usepackage{orcidlink}
\begin{document}

\title{CNN-Transformer Rectified Collaborative Learning for Medical Image Segmentation}

\author{Lanhu Wu \hspace{-1.5mm}$^{~\orcidlink{0009-0006-6420-5971}}$, Miao Zhang \hspace{-1.5mm}$^{~\orcidlink{0000-0002-7972-7047}}$, \IEEEmembership{Member, IEEE}, Yongri Piao \hspace{-1.5mm}$^{~\orcidlink{0000-0002-0860-252X}}$, \IEEEmembership{Member, IEEE}, Zhenyan Yao \hspace{-1.5mm}$^{~\orcidlink{0009-0008-2833-8325}}$,\\Weibing Sun \hspace{-1.5mm}, Feng Tian \hspace{-1.5mm}, and Huchuan Lu \hspace{-1.5mm}$^{~\orcidlink{0000-0002-6668-9758}}$, \IEEEmembership{Fellow, IEEE}
        % <-this % stops a space
\thanks{This work was supported by the National Natural Science Foundation of China under Grant 62172070, Grant 62372080, and Grant 62376050. (Corresponding author: Yongri Piao).}
\thanks{Lanhu Wu, Yongri Piao, Zhenyan Yao, and Huchuan Lu are with
	the School of Information and Communication Engineering, Dalian
	University of Technology, Dalian 116024, China (e-mail: lanhoong0406@gmail.com; yrpiao@dlut.edu.cn;
	yzy@mail.dlut.edu.cn; lhchuan@dlut.edu.cn).}% <-this % stops a space
\thanks{Miao Zhang is with the Key Laboratory for Ubiquitous Network
	and Service Software of Liaoning Province, the DUT-RU International School of Information
	Science and Engineering, Dalian University of Technology,
	Dalian 116024, China (e-mail: miaozhang@dlut.edu.cn).}
\thanks{Weibing Sun and Feng Tian are with the Affiliated Zhongshan Hospital of Dalian University, Dalian 116024, China (e-mail: tianfeng\_dl@sohu.com; massurm@163.com).}}

\markboth{Journal of \LaTeX\ Class Files,~Vol.~14, No.~8, August~2021}%
{Shell \MakeLowercase{\textit{et al.}}: A Sample Article Using IEEEtran.cls for IEEE Journals}


\maketitle

\begin{abstract}
	Automatic and precise medical image segmentation (MIS) is of vital importance for clinical diagnosis and analysis. Current MIS methods mainly rely on the convolutional neural network (CNN) or self-attention mechanism (Transformer) for feature modeling. However, CNN-based methods suffer from the inaccurate localization owing to the limited global dependency while Transformer-based methods always present the coarse boundary for the lack of local emphasis. Although some CNN-Transformer hybrid methods are designed to synthesize the complementary local and global information for better performance, the combination of CNN and Transformer introduces numerous parameters and increases the computation cost. To this end, this paper proposes a CNN-Transformer rectified collaborative learning (CTRCL) framework to learn stronger CNN-based and Transformer-based models for MIS tasks via the bi-directional knowledge transfer between them. Specifically, we propose a rectified logit-wise collaborative learning (RLCL) strategy which introduces the ground truth to adaptively select and rectify the wrong regions in student soft labels for accurate knowledge transfer in the logit space. We also propose a class-aware feature-wise collaborative learning (CFCL) strategy to achieve effective knowledge transfer between CNN-based and Transformer-based models in the feature space by granting their intermediate features the similar capability of category perception. Extensive experiments on three popular MIS benchmarks demonstrate that our CTRCL outperforms most state-of-the-art collaborative learning methods under different evaluation metrics.
\end{abstract}


\section{Introduction}
\IEEEPARstart{M}{edical} image segmentation (MIS) is one of the critical steps in pre-treatment diagnoses, treatment planning, and post-treatment assessments of various diseases. Traditional MIS methods mainly rely on template matching \cite{chen2009automated}, edge detection \cite{yu2006medical}, machine learning \cite{li2004svm}, etc. However, the challenge of selecting discriminating features and appropriate hyper-parameters limits the development of these methods. 
\begin{figure}[t]
	\centering 
	\includegraphics [width=\linewidth] {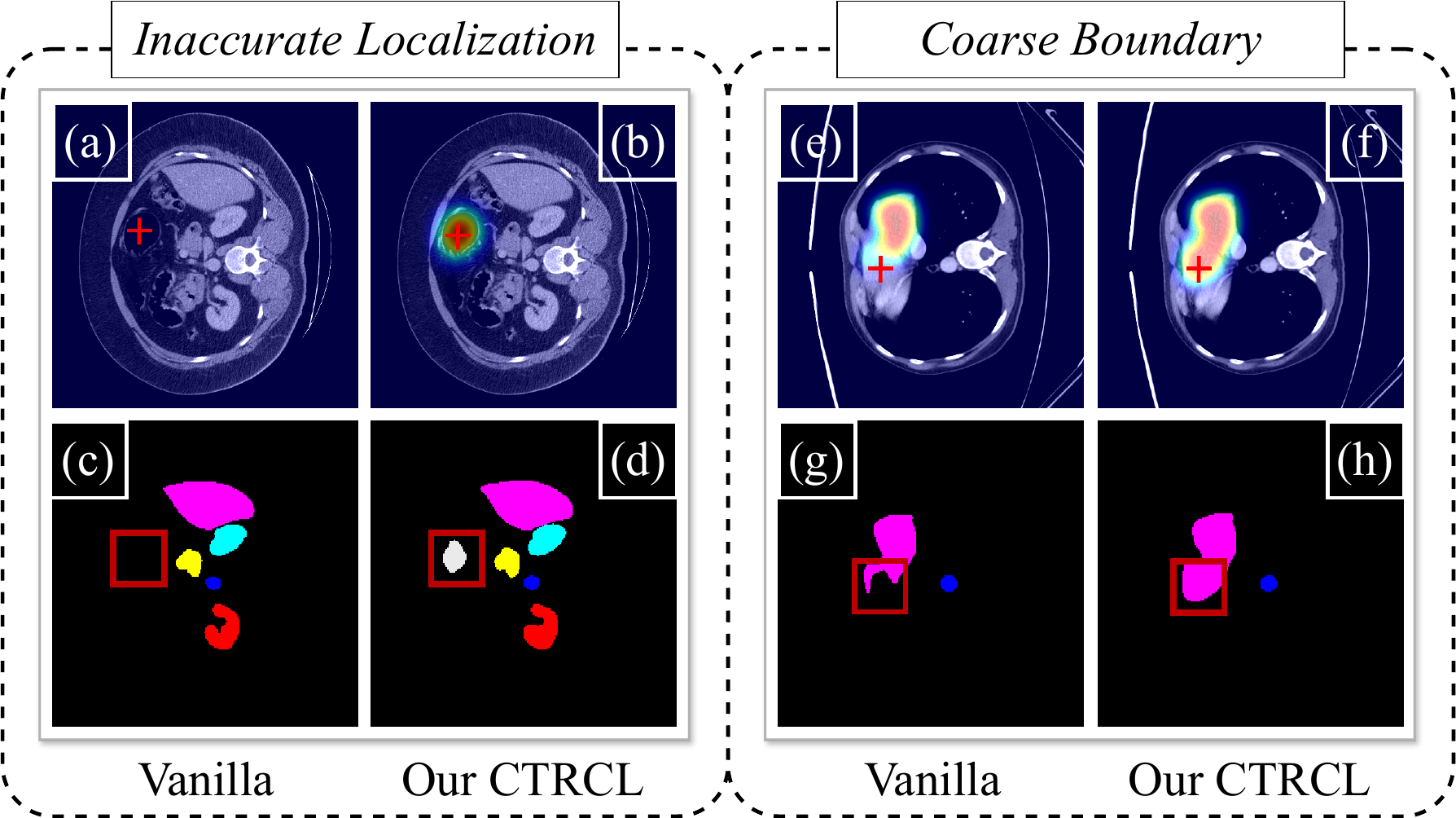}
	\caption{Visualizations of class activation maps generated by Grad-CAM \cite{selvaraju2017grad} and segmentation results of ResNet-50 \cite{he2016deep} (Left) and MiT-B2 \cite{xie2021segformer} (Right). Current CNN-based models suffer from the inaccurate localization (\emph{e.g.}, missing stomach in (a), (c)) while Transformer-based models present the coarse boundary (\emph{e.g.}, incomplete liver in (e), (g)). Our CTRCL improves the performance of the CNN-based model with more accurate location ((b), (d)) and the Transformer-based model with more elaborate boundary ((f), (h)) via collaborative learning between CNN-based and Transformer-based models.}
	\label{fig:motivation}
\end{figure}

In recent years, deep learning technology has been widely applied to MIS and made breakthrough progress. Early deep learning based methods depend on the convolutional neural network (CNN) for hierarchical feature modeling. UNet \cite{ronneberger2015u} produces the high-resolution segmentation map by aggregating multi-stage features with skip connections and integrating them in a top-down manner. Due to the effective encoder-decoder architecture, a few variants of UNet \cite{zhou2018unet++}, \cite{gu2019net}, \cite{feng2020cpfnet}, \cite{wang2020automatic}, \cite{zhou2024uncertainty} have demonstrated the impressive performance in MIS tasks. Despite the effectiveness of CNN-based methods, they exhibit limitations in extracting the global context information due to the confined receptive field of the convolution operation, leading to the inaccurate localization of various organs and lesions (as shown in Fig. \ref{fig:motivation} (a), (c)). To overcome this problem, a series of Transformer-based methods \cite{cao2022swin}, \cite{lin2022ds}, \cite{huang2022missformer}, \cite{li2023erdunet} are proposed for MIS which utilize the self-attention mechanism to model long-range dependencies with dynamic weights and global receptive field. Nevertheless, since each element is calculated and compared with other elements equally in the self-attention process, the local information would be diluted within a global perspective, resulting in the coarse boundary for medical objects (as shown in Fig. \ref{fig:motivation} (e), (g)). Besides, Transformers are data-hungry models while abundant and quality-annotated medical data are difficult to acquire, which limits the potential of these methods. Thus, some methods \cite{chen2021transunet}, \cite{zhang2021transfuse}, \cite{yuan2023effective} design CNN-Transformer hybrid networks to synthesize the complementary local and global information for better performance of MIS. Although these methods achieve significant improvements, the combination of CNN and Transformer introduces numerous parameters and increases the cost of computation. To this end, we ask a question: `why not enable CNN-based and Transformer-based models to learn the complementary global and local information from each other for enhancement of each model?'

Our inspiration is derived from the collaborative learning mechanism that interchanges the knowledge among multiple students synchronously during the end-to-end training process for improvement of each student. However, accurate and effective CNN-Transformer collaborative learning is a challenging task for two reasons. On the one hand, since the CNN-based student and Transformer-based student always suffer from the inaccurate localization and coarse boundary respectively for various MIS tasks, their soft labels (predicted probability) are imprecise in the corresponding areas. Direct collaborative learning in these areas between two students may cause them to converge in wrong directions, thereby degrading their performance. This impact is exacerbated in the initial phase of collaborative learning because both students are unpretrained on the target dataset, leading to severe inaccuracy of their soft labels. On the other hand, CNN and Transformer are heterogeneous networks with substantial intrinsic gaps between their intermediate features, which significantly increases the difficulty of knowledge transfer in the feature space. Forcibly aligning the mismatched features would disrupt their respective characteristics, hence leading to incorrect predictions.

To address these problems, this article proposes a CNN-Transformer rectified collaborative learning (CTRCL) framework to learn stronger CNN-based and Transformer-based models for MIS. Specifically, CTRCL framework incorporates a rectified logit-wise collaborative learning (RLCL) strategy and a class-aware feature-wise collaborative learning (CFCL) strategy to achieve bi-directional knowledge transfer between CNN-based and Transformer-based students in logit and feature spaces, respectively. For accurate logit-wise knowledge transfer, RLCL strategy introduces an adaptive rectification module (ARM) which employs the ground truth to select the wrong regions in student soft labels and to rectify them with dynamic weights. As such, our RLCL is endowed with the ability to adaptively generate high-quality student soft labels for logit-wise mutual learning. For effective feature-wise knowledge transfer, CFCL strategy transforms the student immediate features into class-aware representations via a category perception module (CPM) and aligns them under loss supervision. By granting the student features the similar capability of category perception, CFCL manages to transfer the feature-wise knowledge between heterogeneous networks. To verify the effectiveness of our CTRCL framework, we conduct experiments on three popular MIS benchmarks: Synapse Multi-organ \cite{landman2015miccai}, ACDC \cite{bernard2018deep} and Kvasir-SEG \cite{jha2020kvasir}. The contributions of this work can be summarized as follows:
\begin{itemize}
	\item To our best knowledge, our CTRCL framework makes the first attempt to adopt the collaborative learning mechanism to learn stronger CNN-based and Transformer-based models for MIS tasks via the bi-directional knowledge transfer between them in both logit and feature spaces. 
	
	\item We propose an RLCL strategy for accurate logit-wise knowledge transfer, which introduces the ground truth to adaptively select and rectify the wrong regions in student soft labels with dynamic weights.
	
	\item We propose a CFCL strategy for effective feature-wise knowledge transfer between heterogeneous networks by granting their intermediate features the similar capability of category perception.
	
	\item Our CTRCL framework consistently achieves new state-of-the-art performance on three MIS benchmarks compared with other collaborative learning methods. Particularly, our CTRCL reduces the MAE metric by 42.93\% and 31.23\% for ResNet-50 and MiT-B2 respectively on the Kvasir-SEG \cite{jha2020kvasir} dataset. 
\end{itemize}

%%%%%%%%% -------the overall architecture of our methods--------------
\begin{figure*}[t]
	\centering 
	\includegraphics [width=\linewidth] {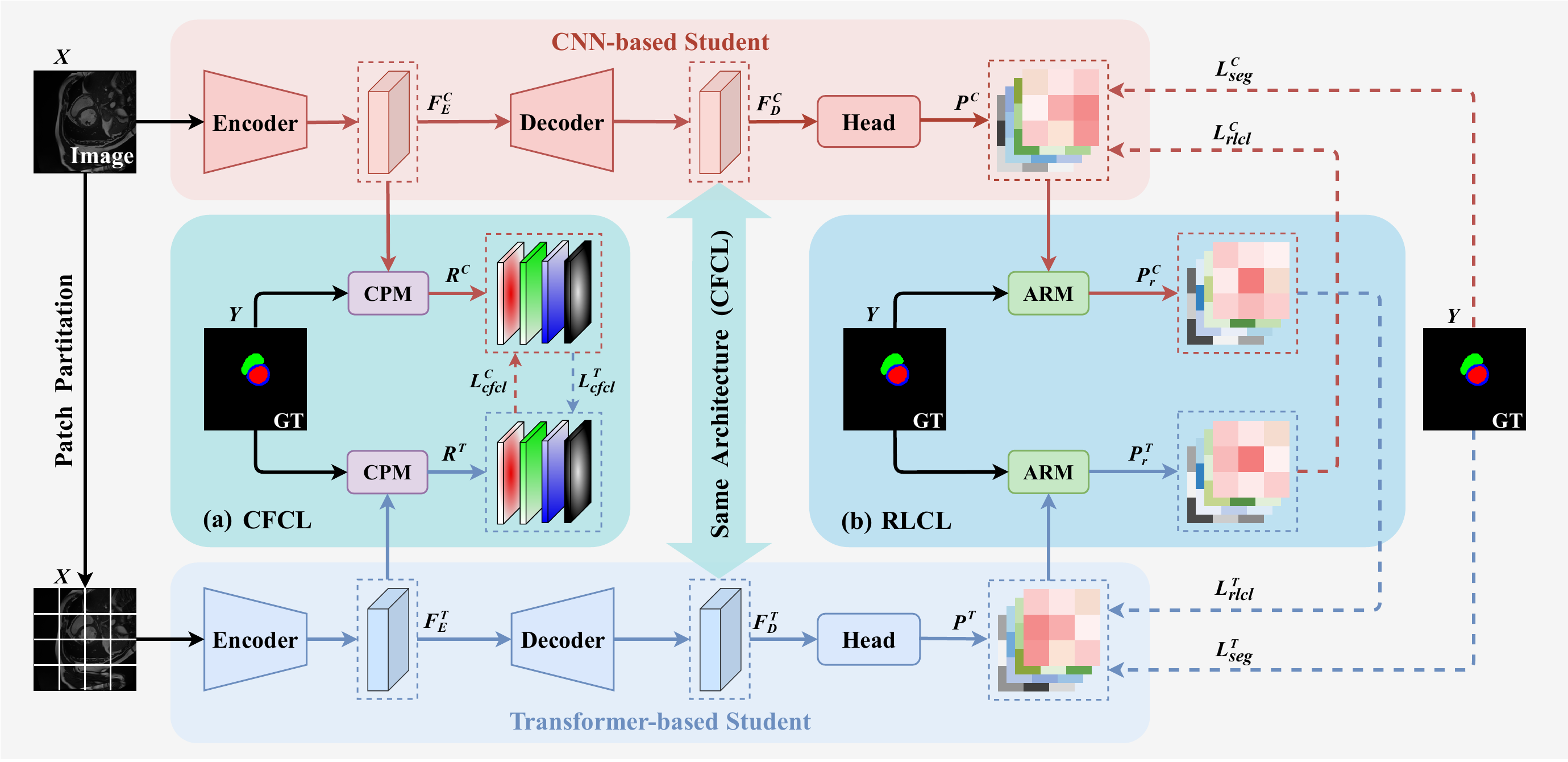}
	%\vspace{-0.4cm}
	\caption{The whole pipeline of our CTRCL framework, containing three parts: CNN-based student, Transformer-based student, and collaborative learning strategies (CFCL and RLCL). (a) Class-aware feature-wise collaborative learning (CFCL) focuses on the effective feature-wise knowledge transfer by encouraging student features to possess similar class-aware representations. (b) Rectified logit-wise collaborative learning (RLCL) aims at the accurate logit-wise knowledge transfer with student soft labels rectified by the ground truth. Please refer to Section \ref{Methodology} for details.}
	\label{fig:overall architecture}
	%\vspace{-0.15cm} 
\end{figure*}

%-------------------------------------------------------------------------

\section{Related Work}

\subsection{Medical Image Segmentation}
Medical image segmentation (MIS) is a dense prediction task that classifies the pixels of organs or lesions in medical images (e.g., CT, MRI, US, endoscopy, etc.). Traditional MIS methods typically rely on hand-crafted features such as edge, shape, texture, etc, which suffer from a high risk of mis-segmentation for the inherent lack of high-level semantics. 

With the development of deep learning technology, CNN-based MIS methods are first proposed and achieve impressive performance. Ronneberger et al. \cite{ronneberger2015u} proposed a symmetric U-shape network to aggregate multi-stage features with skip connections and progressively integrated them for high-resolution segmentation maps. Later, Zhou et al. \cite{zhou2018unet++} designed a series of nested, dense skip connections to reduce the semantic gaps between the feature maps and utilized deep supervision for multi-scale lesions. Jha et al. \cite{jha2019resunet++} introduced the residual connection \cite{he2016deep}, SE \cite{hu2018squeeze} and ASPP\cite{chen2017deeplab} to further improve the performance. However, limited by the receptive field in CNN, these methods are deficient in extracting global context information, leading to inaccurate localization. To this end, some Transformer-based MIS methods are proposed with self-attention mechanism applied for the long-range modeling. Karimi et al. \cite{karimi2021convolution} first presented a convolution-free segmentation model by forwarding flattened image representations to Transformer, whose outputs are then reorganized into 3D tensors to generate segmentation masks. Cao et al. \cite{cao2022swin} utilized the Swin Transformer to construct the encoder and decoder within the U-Net paradigm. Furthermore, Lin et al. \cite{lin2022ds} adopted a dual-scale Transformer encoder to attain coarse–fine-tuning features and employed the self-attention mechanism to establish global dependencies between them. Nonetheless, due to the local dilution in the self-attention process, methods based on pure transformer usually suffer from the coarse boundary in segmentation results. Besides, the excessive data dependency limits the potential of these methods for the insufficient medical data. Thus, some works couple both CNN and Transformer to design hybrid networks for MIS. Chen et al. \cite{chen2021transunet} concatenated a Transformer-based encoder after the CNN-based encoder for global feature extraction. Zhang et al. \cite{zhang2021transfuse} utilized shallow CNN-based encoder and Transformer-based segmentation network in parallel, and fused the features from two branches to jointly make predictions. Zhou et al. \cite{zhou2023nnformer} constructed a CNN-Transformer hybrid backbone, nnFormer, with convolutional and Transformer blocks interleaved to take advantages of each other. Although these methods achieve better performance, the large model size and high computational cost prevent them from application. Different from the aforementioned methods, our CTRCL enables CNN-based model and Transformer-based model to learn from each other to improve the performance of each model. 

\subsection{Collaborative Learning}
Collaborative learning technology derives from the knowledge distillation \cite{hinton2015distilling} (KD) which transfers the knowledge from a cumbersome pre-trained teacher model to a compact student model for model compression. Differently, collaborative learning simultaneously exchanges the knowledge among a series of student models in an end-to-end training procedure to enhance the performance of each model. 

Current collaborative learning methods can be primarily categorized into two types, mutual learning \cite{zhang2018deep}, \cite{zhang2021student}, \cite{kong2022mdflow}, \cite{yang2023online} and ensemble learning \cite{zhu2018knowledge}, \cite{wu2021peer}, \cite{zhao2022distillation}, \cite{su2022deep}. Mutual learning strategy is first time proposed by Zhang et al. \cite{zhang2018deep} to make peer students learn from each other through the Kullback-Leibler (KL) divergence loss between each pair of student logits. Following this work, Chung et al. \cite{chung2020feature} introduced the mutual learning mechanism to feature maps using an adversarial training paradigm. More recently, Zhu et al. \cite{zhu2023good} proposed a bidirectional selective distillation strategy to transfer reliable knowledge between student models in both logit and feature spaces. Ensemble learning constructs a virtual teacher by assembling the outputs of all the students, which are then distilled back to foster each student. Lan et al. \cite{zhu2018knowledge} constructed a multi-branch network and assembled the on-the-fly logit information from the branches to enhance the performance on the target network. Later, Guo et al. \cite{guo2020online} extended the ensemble learning to the outputs of peer students with different architectures to generate high-quality labels for supervision. Furthermore, Kim et al. \cite{kim2021feature} introduced a feature fusion module which integrates the feature representations of sub-networks to construct the teacher classifier whose knowledge is delivered back to each sub-network. Unfortunately, current collaborative learning methods overlook the accuracy of the soft labels in the process of logit-wise knowledge transfer because neither a single student nor an ensemble teacher can completely reflect the distribution of the ground truth. Additionally, feature-wise collaborative learning between heterogeneous networks are not fully explored in previous studies. In contrast, we strive to ensure the accuracy of student soft labels with the help of the ground truth and to achieve effective feature-wise knowledge transfer between heterogeneous networks (CNN and Transformer) via the alignment of class-aware representations.

\section{Proposed Method}
\label{Methodology}

\subsection{Overview}
An overview of our CTRCL framework is depicted in Fig. \ref{fig:overall architecture}, which consists of three components: a CNN-based student $f\left(\theta^{C}\right)$, a Transformer-based student $f\left(\theta^{T}\right)$, and the proposed collaborative learning strategies (RLCL and CFCL). Given an input image set $\mathcal{X}$ and its label set $\mathcal{Y}$, our objective is to enable $f\left ( X;\theta ^{C}  \right )$ and $f\left ( X;\theta ^{T}  \right )$ to learn collaboratively to assign the pixel-wise label $l \in 1,...,L$ in image $X \in \mathcal{X}\left(X \in \mathbb{R}^{H\times W\times D }\right)$ more accurately than the student itself, where $H$, $W$ and $D$ are the height, width and depth of $X$, $L$ is the number of categories. To achieve this goal, given a specific input $X$, we first attain the segmentation predictions ($P^{C}$ and $P^{T}$) and feature representations ($F^{C}$ and $F^{T}$) from the two students $f\left ( X;\theta ^{C}  \right )$ and $f\left ( X;\theta ^{T}  \right )$, formulated as: 
\begin{equation} 
	\begin{aligned}
		\left ( P^{C},F^{C} \right ) &=f\left ( X;\theta ^{C}  \right ) ,\\
		\left ( P^{T},F^{T} \right ) &=f\left ( X;\theta ^{T}  \right ).
	\end{aligned}			
\end{equation}
Then the pixel-wise segmentation loss ($\mathcal{L}_{seg}$) is based on the cross entropy $\mathrm{CE\left ( \cdot \right ) } $ with the ground truth $Y$:
\begin{equation} 
	\begin{aligned}
		{\mathcal{L}}_{seg}^{C} & = \frac{1}{H\times W} \sum_{h= 1}^{H}  \sum_{w= 1}^{W} \mathrm{CE}\left (  P^{C}_{\left ( h,w \right ) }   , Y_{\left ( h,w \right )} \right )  ,    \\
		{\mathcal{L}}_{seg}^{T} & = \frac{1}{H\times W} \sum_{h= 1}^{H}  \sum_{w= 1}^{W} \mathrm{CE}\left (  P^{T}_{\left ( h,w \right ) }   , Y_{\left ( h,w \right )} \right ),
	\end{aligned}
\end{equation}
where ${\left ( h,w \right )}$ indexes the spatial location for each pixel. Meanwhile, we propose a rectified logit-wise collaborative learning (RLCL) strategy and a class-aware feature-wise collaborative learning (CFCL) strategy to achieve bi-directional knowledge transfer between CNN-based and Transformer-based students in the logit and feature spaces, respectively. The details of RLCL and CFCL strategies are described as follows.

\subsection{Rectified Logit-wise Collaborative Learning}
It is a fact that neither a CNN-based nor a Transformer-based student can completely mimic the ground truth, inevitably leading to mis-categorized pixels in their soft labels, which is adverse to the logit-wise collaborative learning between them. Consequently, we propose a rectified logit-wise collaborative learning (RLCL) strategy which adopts the ground truth to adaptively rectify the wrong regions in student soft labels for accurate logit-wise knowledge transfer between CNN-based and Transformer-based students.   

As shown in Fig. \ref{fig:overall architecture} (b), the RLCL strategy introduces an adaptive rectification module (ARM) to rectify the segmentation predictions of both students ($P^{C}$ and $P^{T}$) and obtain rectified soft labels ($P^{C}_{r}$ and $P^{T}_{r}$) under the guidance of the ground truth $Y$, which can be formulated as:
\begin{equation} 
	\begin{aligned}
		{P}_{r}^{C}  &= \mathrm{ARM} \left({P}^{C}, Y\right) , \\
		{P}_{r}^{T}  &= \mathrm{ARM} \left({P}^{T}, Y\right).
	\end{aligned} 	
\end{equation}

\begin{figure}[t]
	\centering 
	\includegraphics [width=\linewidth] {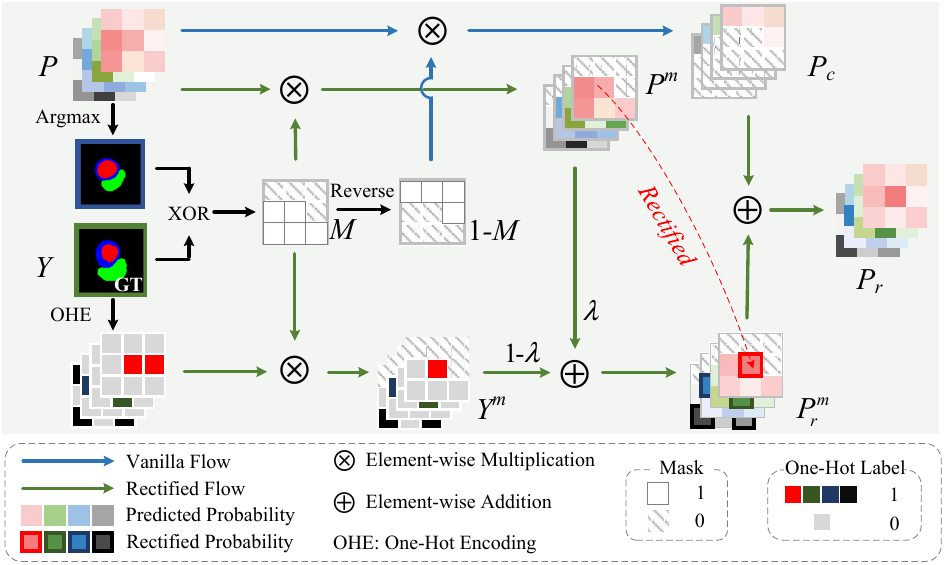}
	%\vspace{-0.4cm}
	\caption{Illustration of the adaptive rectification module (ARM).}
	\label{fig:ARM}
	%\vspace{-0.1cm}
\end{figure}

The illustration of ARM is provided in Fig. \ref{fig:ARM}. Given the segmentation prediction of a student as $P$, we first attain its pseudo label via the Argmax function and compare it with the ground truth $Y$ by the XOR operation to produce an error mask $M$ as follows:
\begin{equation}
	M= \mathrm{XOR}\left( \mathrm{Argmax}\left ( P \right ), Y\right) .
\end{equation}
Then we restore the one-hot label from the ground truth $Y$ and utilize the error mask $M$ to select the mis-segmented region $Y^{m}$ and $P^{m}$ from $Y$ and $P$, respectively. After that, $P^{m}$ is rectified to $P_{r}^{m}$ by the combination of $P^{m}$ and $Y^{m}$. The above process can be written as: 
\begin{equation}
	\begin{aligned}
		Y^{m} &= \mathrm{OHE}\left ( Y \right ) \cdot M, \\
		P^{m} &= P \cdot M,  \\
		P^{m}_{r} &= \lambda \cdot P^{m} + \left( 1 - \lambda\right) \cdot Y^{m} ,
	\end{aligned}
\end{equation}
%\begin{equation}
%	\begin{aligned}
%		\left\lbrace   Y^{m}, \, P^{m}\right\rbrace     &= \left\lbrace   \mathrm{OHE}\left ( Y \right ) \cdot M, \, P \cdot M\right\rbrace  ,  \\
%		P^{m}_{r} &= \lambda \cdot P^{m} + \left( 1 - \lambda\right) \cdot Y^{m} ,
%	\end{aligned}
%\end{equation}
where $\mathrm{OHE}\left(\cdot\right)$ denotes the one-hot encoding and $\lambda$ is a matrix of dynamic weights for rectification demonstrated afterwards. Finally, we joint the rectified region $P_{r}^{m}$ and the correct region $P_{c}$ to generate the rectified soft label $P_{r}$, formulated as:
\begin{equation}
	\begin{aligned}
		P_{c} &= P \cdot (1-M), \\
		P_{r} &= P_{r}^{m} + P_{c}.
	\end{aligned}	
\end{equation}

Now, we describe the computation of $\lambda$ and the process of rectification in detail. As a whole, the $\lambda$ consists of an alignment factor ${\lambda}^{a}$, a similarity-based decay factor ${\lambda}^{s}$ and a certainty-based decay factor ${\lambda}^{c}$, formulated as:
\begin{equation}
	\lambda = {\lambda}^{a} \cdot {\lambda}^{s} \cdot {\lambda}^{c}.
\end{equation}
\subsubsection{Alignment Factor (${\lambda}^{a}$)}
We initially introduce the one-hot label in $Y^{m}$ to align the probabilities of mis-categorized class and truth class in the prediction $P^{m}$. Given the probabilities of mis-categorized class and truth class in $P^{m}$ as $p^{mis}$ and $p^{tru}$, the alignment process can be formulated as:
\begin{equation}
\label{eq:eq_lambda_b}
{\lambda}^{a} \cdot p^{mis} + \left( 1 - {\lambda}^{a} \right)\cdot v^{mis} = {\lambda}^{a} \cdot p^{tru} + \left( 1 - {\lambda}^{a} \right) \cdot v^{tru}
\end{equation}
where $v^{mis}$ and $v^{tru}$ are the values of mis-categorized class and truth class in $Y^{m}$, namely 0 and 1 respectively. Thus, the alignment factor $\lambda^{a}$ could be obtained as follows:
\begin{equation}
{\lambda}^{a} = {\left( 1+p^{mis}-p^{tru}\right) }^{-1}.
\end{equation}
Let $\lambda:=\lambda^a$, the mis-categorized class shares the same probability with the truth class in $P^m_r$. Furthermore, ARM aims to modify the mis-category prediction, which means a larger probability on truth class than mis-category class. Namely, the equality relationship in Eq.~(\ref{eq:eq_lambda_b}) should be converted to `$<$'. To achieve this, we additionally introduce two decay factors ($\lambda^s$ and $\lambda^c$) on the basis of $\lambda^a$.

\subsubsection{Similarity-based Decay Factor (${\lambda}^{s}$)}
${\lambda}^{s}$ is a decay factor to decrease the proportion of prediction for rectification, which is positively correlated to the similarity between the prediction and the one-hot label. We adopt the cross entropy $\mathrm{CE\left ( \cdot \right ) } $ to measure the similarity and formulate the ${\lambda}^{s}$ as follows:
\begin{equation}
	{\lambda}^{s} = \mathrm{exp}\left( - \mathrm{CE}\left( P^{m},Y^{m}  \right)  \right).
\end{equation}

\subsubsection{Certainty-based Decay Factor (${\lambda}^{c}$)}
Similarly, ${\lambda}^{c}$ is also a decay factor for the rectification process, positively correlated to the certainty of the prediction. We adopt the entropy $\mathrm{S}\left( \cdot\right) $ to evaluate the certainty of predictions and calculate the ${\lambda}^{c}$ by the following formula:
\begin{equation}
	{\lambda}^{c} = \mathrm{exp}\left( - \mathrm{S}\left( P^{m}\right) \right) .
\end{equation}

%Meanwhile, since the probability of the annotated class is 1 and ones of other classes are 0 in $Y^{m}$, the quantitative relationship among other classes in $P^{m}$ are unchanged in the rectification process, which reserves the structural knowledge as much as possible. 
In this way, the ARM adaptively select and rectify the wrong regions in student soft labels under the guidance of ground truth with dynamic weights to generate accurate soft labels for logit-wise collaborative learning. 

With the rectified soft labels ($P^{C}_{r}$ and $P^{T}_{r}$) obtained, we follow the previous approaches to adopt the KL divergence $\mathrm{KL\left(\cdot\right)}$ for logit-wise collaborative learning between CNN-based and Transformer-based students:
\begin{equation} 
	\begin{aligned}
		{\mathcal{L}}_{rlcl}^{C} & = \frac{1}{H\times W} \sum_{h= 1}^{H}  \sum_{w= 1}^{W} \mathrm{KL}\left ( P^{C}_{\left ( h,w \right )} || {P^{T}_{r}}_{\left ( h,w \right )} \right )  ,    \\
		{\mathcal{L}}_{rlcl}^{T} & = \frac{1}{H\times W} \sum_{h= 1}^{H}  \sum_{w= 1}^{W} \mathrm{KL}\left ( P^{T}_{\left ( h,w \right )} || {P^{C}_{r}}_{\left ( h,w \right )} \right ).
	\end{aligned}
\end{equation}

\subsection{Class-aware Feature-wise Collaborative Learning}
The distinct modeling approaches of CNN and Transformer result in intrinsic feature gaps between them. To this end, we propose a class-aware feature-wise collaborative learning (CFCL) strategy to achieve effective feature-wise knowledge transfer between heterogeneous CNN-based and Transformer-based students by granting their immediate features the similar capability of category perception. 

\begin{figure}[b]
	\centering 
	\includegraphics [width=\linewidth] {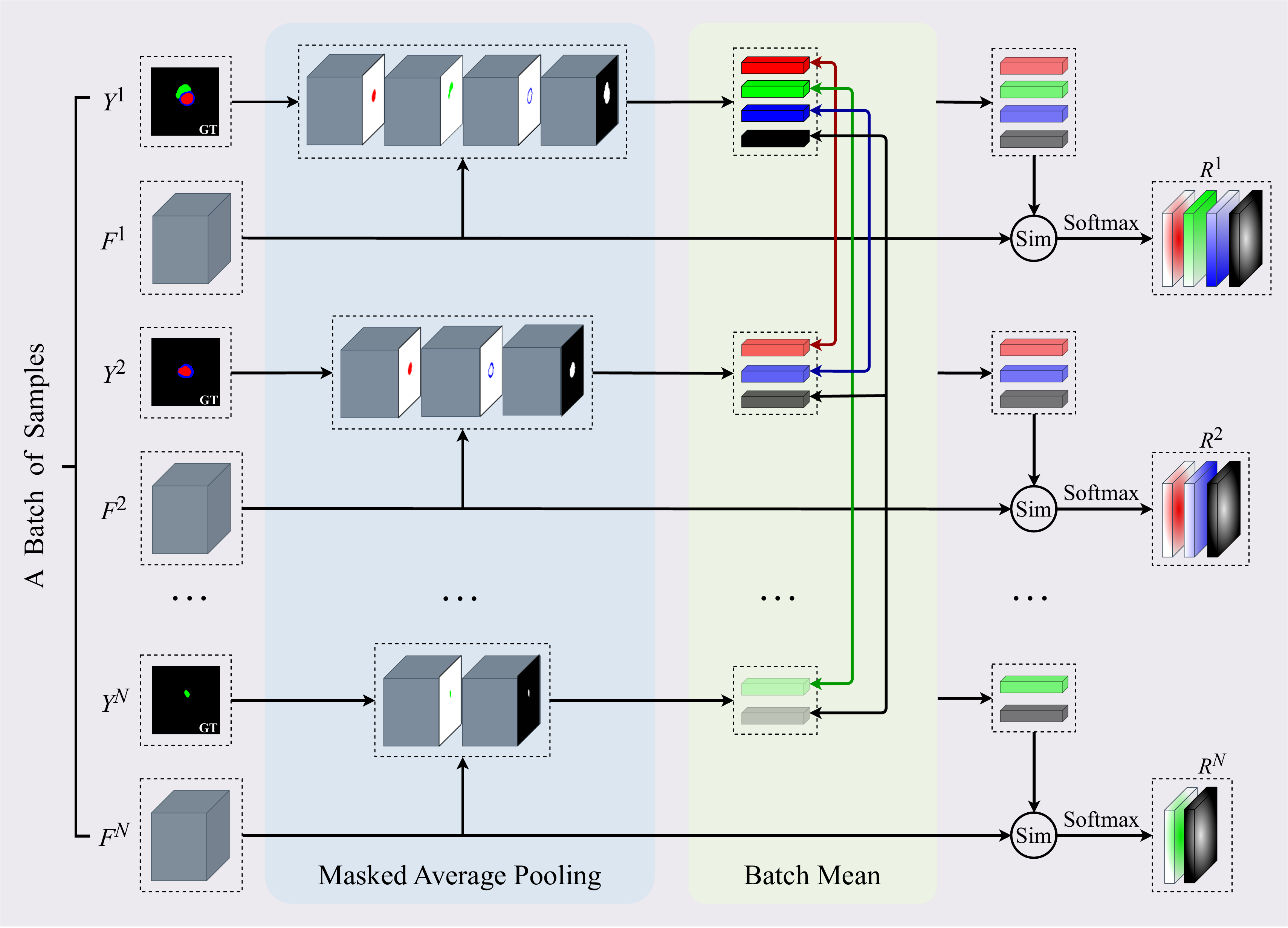}
	%\vspace{-0.4cm}
	\caption{Illustration of the category perception module (CPM).}
	\label{fig:CPM}
	%\vspace{-0.1cm}
\end{figure}

As shown in Fig. \ref{fig:overall architecture} (a), the CFCL strategy introduces a category perception module (CPM) to extract the class-aware representations ($R^{C}$ and $R^{T}$) for the categories in ground truth $Y$ from immediate features ($F^{C}$ and $F^{T}$), which can be formulated as:
\begin{equation} 
	\begin{aligned}
		R^{C}  &= \mathrm{CPM} \left({F}^{C}, Y\right) , \\
		R^{T}  &= \mathrm{CPM} \left({F}^{T}, Y\right).
	\end{aligned} 	
\end{equation}

The structure of the CPM is illustrated in Fig. \ref{fig:CPM}. Given a batch of input images $\left\lbrace X^n\right\rbrace ^N_{n=1}$ where $N$ denotes the batch size, and their corresponding ground truths $\left\lbrace Y^n\right\rbrace ^N_{n=1}$ with $K$ classes (including background) in total, let $F^n \in \mathbb{R}^{\hat{H}\times \hat{W}\times \hat{D} }$ be the feature map of a student network, where $\hat{H}$, $\hat{W}$ and $\hat{D}$ are the height, width and depth of $F^n$. The prototype of class $l \in \mathcal{K}$ (with $\left | \mathcal{K} \right |  = K$) is generated via the masked average pooling and batch mean:
\begin{equation} 	
	{p}_{l} = \frac{1}{N_{l}}\sum_{n}{\frac{ {\textstyle \sum_{\hat{h},\hat{w}}} F^{n}_{\left ( \hat{h},\hat{w} \right ) }\mathbbm{1}\left [ Y^{n}_{\left ( \hat{h},\hat{w} \right ) } =l \right ]}{ {\textstyle \sum_{\hat{h},\hat{w}}} \mathbbm{1}\left [ Y^{n}_{\left ( \hat{h},\hat{w} \right ) } =l \right ]+\epsilon}} ,   
\end{equation}
where the ground truth $Y^{n}$ is down-sampled to match the spatial size of the feature map $F^{n}$ with the nearest interpolation; $ ( \hat{h},\hat{w}  )$ indexes the spatial locations for each pixel; $\mathbbm{1}\left [ \cdot \right ]$ represents the indicator function that returns value 1 when the argument is true or 0 otherwise; $\epsilon$ is an infinitesimal to cope with the absence of some classes in a batch; and $N_{l}$ is the number of images that include the $l$th class. 

Since the prototype actually denotes the channel-wise summary of a class, we can obtain the pixel-wise category perception by establishing the relationship between the feature map and the prototype. To achieve this objective, we adopt a non-parametric metric learning mechanism. Concretely, we first calculate the distances between the feature vector at each spatial location and all prototypes. Then we apply a softmax operation over the distances for the class-aware representation $R$. Let $\mathcal{P}^{n}=\left \{ p_{l}\mid  l\in  \mathcal{K}^{n} , \, \left | \mathcal{K}^{n} \right |  = K^n \right \} $ be the prototypes for the $n$th sample where $K^n$ is the number of classes in the $n$th sample, for each $p_{i} \in \mathcal{P}$, we have:
\begin{equation} 	
	{R^n_i}_{\left ( \hat{h}, \hat{w}  \right ) } = \frac{\mathrm{exp}\left ( - \alpha d\left ( F^n_{\left ( \hat{h}, \hat{w}  \right ) }  ,p_{i} \right )  \right ) }{ {\textstyle \sum_{p_{i} \in \mathcal{P}^{n} }} \mathrm{exp}\left ( - \alpha d\left ( F^n_{\left ( \hat{h}, \hat{w}  \right ) }  ,p_{i} \right )  \right )  } , 
\end{equation}
where $d\left ( \cdot  \right ) $ adopts the cosine distance, and multiplier $\alpha$ is a scaling factor fixed at 20 recommended by \cite{wang2019panet}. 

To achieve the class-aware feature-wise collaborative learning between CNN-based and Transformer-based students, we propose to minimize the loss function as follows:
\begin{equation} 
	\begin{aligned}
		{\mathcal{L}}_{cfcl}^{C} & = \frac{1}{\hat{H}\times \hat{W}} \sum_{\hat{h}= 1}^{\hat{H}}  \sum_{\hat{w}= 1}^{\hat{W}} \mathrm{KL}\left ( {R}^{C}_{\left ( \hat{h},\hat{w} \right )} || {R}^{T}_{\left ( \hat{h},\hat{w} \right )} \right )  ,    \\
		{\mathcal{L}}_{cfcl}^{T} & = \frac{1}{\hat{H}\times \hat{W}} \sum_{\hat{h}= 1}^{\hat{H}}  \sum_{\hat{w}= 1}^{\hat{W}} \mathrm{KL}\left ( {R}^{T}_{\left ( \hat{h},\hat{w} \right )} || {R}^{C}_{\left ( \hat{h},\hat{w} \right )} \right ).
	\end{aligned}
\end{equation}

Noteworthily, we perform the bi-directional CFCL on the output features of both encoder and decoder, i.e., $\left(F^{C}_{E}, F^{T}_{E}\right)$ and $\left(F^{C}_{D}, F^{T}_{D}\right)$, to exchange the information of high-level semantics as well as the low-level details between CNN student and Transformer student, respectively.

\subsection{Optimization}
The overall loss functions of the CTRCL framework for CNN-based and Transformer-based students are given as:
\begin{equation} 
	\begin{aligned}
		{\mathcal{L}}^{C} & =  {\mathcal{L}}_{seg}^{C} + \beta{\mathcal{L}}_{rlcl}^{C} + \gamma_{1} {\mathcal{L}}_{cfcl}^{E,C} + \gamma_{2} {\mathcal{L}}_{cfcl}^{D,C},    \\
		{\mathcal{L}}^{T} & =  {\mathcal{L}}_{seg}^{T} + \beta{\mathcal{L}}_{rlcl}^{T} + \gamma_{1} {\mathcal{L}}_{cfcl}^{E,T} + \gamma_{2} {\mathcal{L}}_{cfcl}^{D,T},
	\end{aligned}
\end{equation}
where $\mathcal{L}_{seg}$ is the segmentation loss between the student predictions and the ground truth; ${\mathcal{L}}_{rlcl}$ is the rectified logit-wise collaborative learning loss between the student predictions; ${\mathcal{L}}_{cfcl}^{E}$ and ${\mathcal{L}}_{cfcl}^{D}$ denote the class-aware feature-wise collaborative learning loss between the student output features of encoder and decoder, i.e., $\left(F^{C}_{E}, F^{T}_{E}\right)$ and $\left(F^{C}_{D}, F^{T}_{D}\right)$ respectively; $\beta$, $\gamma_{1}$ and $\gamma_{2}$ are hyperparameters to trade off these loss terms. The optimization details are summarized in Algorithm~\ref{algorithm_1}. In test, we deploy both CNN-based and Transformer-based students because our target is to learn stronger CNN-based and Transformer-based models for MIS via the collaborative learning mechanism.
\begin{algorithm}[t]
	\caption{Optimization of CTRCL}
	\label{algorithm_1}
	\begin{algorithmic}[1] % The number [1] tells LaTeX to number each line
		\Require Image set $\mathcal{X}$, label set $\mathcal{Y}$, CNN-based student $f(\theta^C)$ and Transformer-based student $f(\theta^V)$;
		\Ensure Two trained student models;
		\State \textbf{Initialization:} Initialize $\theta^C$ and $\theta^V$;
		\Repeat
		\State Input a batch of images $\{X^n, Y^n\}_{n=1}^N$;		
		\State Attain $\left( P^C,P^T\right) $ and $\left( F^C,F^T\right) $ with Eq. (1);
		\State Compute ${\mathcal{L}}_{seg}^{C}$ and ${\mathcal{L}}_{seg}^{T}$ with Eq. (2);
		\State Select $P^{m}$ in $P$ with Eq. (4);
		\State Compute $\lambda^a$, $\lambda^s$ and $\lambda^c$ with Eq. (9) (10) (11) and obtain $\lambda$ with Eq. (7);  
		\State Attain $P_r$ with Eq. (5) and (6);
		\State Compute ${\mathcal{L}}_{rlcl}^{C}$ and ${\mathcal{L}}_{rlcl}^{T}$ with Eq. (12);
		\State Compute the prototype $p$ with Eq. (14);
		\State Attain the class-aware representation $R$ with Eq. (15);
		\State Compute ${\mathcal{L}}_{cfcl}^{C}$ and ${\mathcal{L}}_{cfcl}^{T}$ with Eq. (16);
		\State Compute ${\mathcal{L}}^{C}$ and ${\mathcal{L}}^{T}$ with Eq. (17);
		\State Update $\theta^C$ and $\theta^V$ respectively;	
		\Until {maximum iterations}.
		\State \textbf{return} $\theta^C$ and $\theta^V$.
		\State \textbf{End}.
	\end{algorithmic}
\end{algorithm}

\section{Experiments}
\subsection{Datasets}
\subsubsection{Synapse Multi-organ}
The Synapse \cite{landman2015miccai} dataset contains 30 abdominal CT scans with 3779 axial contrast-enhanced abdominal CT images including 8 anatomical structures, i.e., aorta, gallbladder (GB), left kidney (KL), right kidney (KR), liver, pancreas (PC), spleen (SP), and stomach (SM). Each CT scan consists of 85-198 slices of 512 $\times$ 512 pixels, with a voxel spatial resolution of [0:54-0:54] $\times$ [0:98-0:98] $\times$ [2:5-5:0] $\mathrm{mm}^3$. Following \cite{cao2022swin}, \cite{chen2021transunet}, we divide the dataset into 18 scans (2211 axial slices) for training, and 12 scans (1568 axial slices) for testing.

\subsubsection{Automatic Cardiac Diagnosis Challenge}
The ACDC \cite{bernard2018deep} dataset contains 150 cardiac MRI scans which consist of three organs, right ventricle (RV), myocardium (Myo), and left ventricle (LV). Cine MR images were acquired in breath hold, and a series of short-axis slices cover the heart from the base to the apex of the left ventricle, with a slice thickness of 5 to 8 $\mathrm{mm}$. The short-axis in-plane spatial resolution goes from 0.83 to 1.75 $\mathrm{mm}^2$/pixel. We follow the official division protocol for experiments, i.e., 100 cases (1902 axial slices) for training and 50 cases (1076 axial slices) for testing.

\subsubsection{Kvasir-SEG} 
The Kvasir \cite{jha2020kvasir} dataset is the most commonly used endoscopic image dataset in the field of polyp segmentation, collected by Vestre Viken Health Trust in Norway from inside the gastrointestinal (GI) tract. It contains 1000 images with different resolutions from 720 $\times$ 576 to 1920 $\times$ 1072 pixels related to endoscopic polyp removal, which can be used for computer-aided gastrointestinal lesion segmentation. Consistent with \cite{fan2020pranet}, \cite{shi2022polyp}, we divide the dataset into 900 images for training and 100 images for testing.
\begin{table*}[t]
	%\small
	%\belowrulesep=0pt
	%\aboverulesep=0pt
	%\vspace{-0.35cm}
	\caption{Quantitative Comparisons of DSC ($\%$), JAC ($\%$) and HSD ($\mathrm{mm}$) Scores on Synapse Multi-organ Dataset and ACDC Dataset. $\uparrow$($\downarrow$) Denotes Higher the Better (Lower the Better). The Best Results are Shown in \textbf{Boldface}. Vanilla Means no Collaborative Learning. $\Delta(\%)$ Denotes the Improvement Rate Comparing Our Method with Vanilla.}
	\centering
	\label{table:synapse and ACDC}
	\scalebox{1}{
		\begin{tabular}{clcccccccccccc}
			\toprule
			\multirow{2}{*}{Dataset}     &\multirow{2}{*}{Method}    & \multicolumn{3}{c}{MobileNetV2}	&\multicolumn{3}{c}{MiT-B0}   & \multicolumn{3}{c}{ResNet-50}	&\multicolumn{3}{c}{MiT-B2}	  \\ 
			\cmidrule(r){3-5}       \cmidrule(r){6-8}  \cmidrule(r){9-11}   \cmidrule(r){12-14}
			\multicolumn{2}{c}{~}                                                                                     &DSC$\uparrow$  &JAC$\uparrow$  &HSD$\downarrow$     &DSC$\uparrow$  &JAC$\uparrow$  &HSD$\downarrow$    &DSC$\uparrow$  &JAC$\uparrow$  &HSD$\downarrow$          &DSC$\uparrow$  &JAC$\uparrow$  &HSD$\downarrow$    \\ 		
			\midrule
			\multirow{9}{*}{Synapse} &Vanilla                  &73.35     &62.69           &25.31   &75.68      &65.67     &19.39    &75.62           &65.67    &30.11     &77.85	  &67.64    	    &26.52\\
			&DML \cite{zhang2018deep}                        &73.97     &63.40           &19.84   &75.94      &66.05     &21.82    &76.33           &66.21    &27.45     &78.09	  &68.43    	    &25.13\\
			&AFD \cite{chung2020feature}                                  &74.38     &63.86           &18.86   &76.46      &66.29     &\textbf{12.45}    &76.69           &66.78    &26.12    &78.52	  &68.69    	    &23.62\\
			&CTCL \cite{zhu2023good}                                       &75.01     &64.64           &22.04   &77.00      &67.18     &17.23    &77.39           &67.52    &23.73     &79.44	  &70.17    	    &21.57\\
			&KDCL \cite{guo2020online}                 &74.12     &63.25           &18.05   &76.32      &66.02     &18.61     &76.55                 &66.36    &26.87     &78.25	  &68.56    	    &24.63\\
			&FFL \cite{kim2021feature}                  &74.58     &64.40           &20.21   &76.79      &66.87     &15.95    &76.92                &66.97    &24.34    &78.86	  &69.31    	    &21.56\\
			&AFID \cite{su2021attention}                  &74.91     &64.36           &20.85   &77.16      &67.34     &15.05    &77.16                 &67.24    &23.96    &79.09	  &69.62    	    &22.13\\
			\cmidrule(r){2-14}
			
			&CTRCL (Ours)                  &\textbf{76.09}     &\textbf{65.77}    &\textbf{17.38}   &\textbf{78.03}      &\textbf{68.33}     &15.80    &\textbf{78.64}    &\textbf{69.50}    &\textbf{21.88}     &\textbf{80.81}	  &\textbf{71.91}   	    &\textbf{19.71}\\
			&$\Delta(\%)$                  &3.74     &4.91           &31.33   &3.11      &4.05     &18.51    &3.99                &5.83    &27.33     &3.80	  &6.31    	    &25.68 \\
			
			\midrule
			\multirow{9}{*}{ACDC} &Vanilla                  &86.48     &76.88           &1.96   &88.09      &79.40     &1.33    &88.39                 &79.82    &1.85     &89.22	  &81.12    	    &1.46\\
			&DML \cite{zhang2018deep}                  &86.95     &77.68           &1.48   &87.87      &79.03     &1.67    &88.94                 &80.64    &1.25     &89.12	  &80.99    	    &1.55\\
			&AFD \cite{chung2020feature}                  &87.42     &78.42           &1.35   &88.37      &79.87     &1.28    &89.30                &81.22    &1.21    &89.44	  &81.42    	    &1.27\\
			&CTCL \cite{zhu2023good}                  &88.21     &79.60           &1.37   &88.98      &80.80     &1.22    &89.92                 &82.18    &1.19     &90.33	  &82.85    	    &1.24\\
			&KDCL \cite{guo2020online}                  &87.21     &78.06           &1.58   &88.26      &79.68     &1.52    &89.25                 &81.17    &1.43     &89.56	  &81.65    	    &1.26\\
			&FFL \cite{kim2021feature}                  &87.78     &78.98           &1.28   &88.64      &80.21     &1.52    &89.62                 &81.75    &1.24    &89.75	  &81.94    	    &1.21\\
			&AFID \cite{su2021attention}                  &87.92     &79.09           &1.33   &88.85      &80.57     &1.73    &89.76                 &81.89    &\textbf{1.18}    &89.98	  &82.31    	    &1.23\\
			\cmidrule(r){2-14}
			
			&CTRCL (Ours)             &\textbf{89.01}     &\textbf{80.80}           &\textbf{1.23}   &\textbf{89.84}      &\textbf{82.10}     &\textbf{1.17}    &\textbf{90.52}        &\textbf{83.19}    &1.20     &\textbf{91.19}	  &\textbf{84.25}    	    &\textbf{1.11}\\
			&$\Delta(\%)$                  &2.93     &5.10           &37.24   &1.99      &3.40     &12.03    &2.41                 &4.22    &35.14     &2.21	  &3.86    	    &23.97\\
			
			\bottomrule
	\end{tabular}}
\end{table*}

\begin{table*}[t]
	%\small
	%\belowrulesep=0pt
	%\aboverulesep=0pt
	%\vspace{-0.35cm}
	\caption{Quantitative Comparisons of DSC ($\%$), JAC ($\%$) and MAE ($\%$) Scores on Kvasir Dataset. $\uparrow$($\downarrow$) Denotes Higher the Better (Lower the Better). The Best Results are Shown in \textbf{Boldface}. Vanilla Means no Collaborative Learning . $\Delta(\%)$ Denotes the Improvement Rate Comparing Our Method with Vanilla.}
	\centering
	\label{table:polyp}
	\scalebox{1}{
		\begin{tabular}{clcccccccccccc}
			\toprule
			\multirow{2}{*}{Dataset}     &\multirow{2}{*}{Method}    & \multicolumn{3}{c}{MobileNetV2}	&\multicolumn{3}{c}{MiT-B0}   & \multicolumn{3}{c}{ResNet-50}	&\multicolumn{3}{c}{MiT-B2}	  \\ 
			\cmidrule(r){3-5}       \cmidrule(r){6-8}  \cmidrule(r){9-11}   \cmidrule(r){12-14}
			\multicolumn{2}{c}{~}         &DSC$\uparrow$  &JAC$\uparrow$  &MAE$\downarrow$     &DSC$\uparrow$  &JAC$\uparrow$  &MAE$\downarrow$    &DSC$\uparrow$  &JAC$\uparrow$  &MAE$\downarrow$          &DSC$\uparrow$  &JAC$\uparrow$ &MAE$\downarrow$\\ 		
			\midrule
			\multirow{9}{*}{Kvasir} &Vanilla                  &86.39     &79.64           &3.68   &89.17      &83.64     &3.24    &87.23         &81.12    &3.82     &89.96	  &85.02    	    &2.85\\
			&DML \cite{zhang2018deep}                        &87.26     &81.20           &3.52   &88.94      &83.01     &3.25    &88.43        &82.51    &4.10     &89.72	  &84.39    	    &2.88\\
			&AFD \cite{chung2020feature}                                  &87.77     &81.51           &3.15   &89.64      &83.87     &2.88    &89.22           &83.13    &3.14    &90.51	  &85.24    	    &2.83\\
			&CTCL \cite{zhu2023good}                                       &88.58     &82.82           &2.94   &90.78      &85.37     &2.64    &90.17           &84.86    &2.38     &91.87	  &86.97    	    &2.36\\
			&KDCL \cite{guo2020online}                  &87.52     &81.73           &3.38   &89.40      &83.50     &3.03     &88.91                 &83.04    &3.48    &90.37	  &85.09    	    &2.66\\
			&FFL \cite{kim2021feature}                  &87.94     &81.97           &3.17   &89.97      &84.14     &2.88    &89.57                &84.00    &2.61    &90.92	  &85.86    	    &2.44\\
			&AFID \cite{su2021attention}                  &88.18     &82.47           &3.13   &90.39      &84.93     &2.82    &89.95                 &84.51    &2.58    &91.30	  &86.28    	    &2.53\\
			\cmidrule(r){2-14}
			
			&CTRCL (Ours)                  &\textbf{89.66}     &\textbf{84.34}    &\textbf{2.60}   &\textbf{91.81}      &\textbf{86.60}     &\textbf{2.50}    &\textbf{91.68}    &\textbf{86.55}    &\textbf{2.18}     &\textbf{93.12}	  &\textbf{88.24}   	    &\textbf{1.96}\\
			&$\Delta(\%)$                  &3.79     &5.90           &29.35   &2.96      &3.54     &22.84    &5.10                &6.69    &42.93     &3.51	  &3.79    	    &31.23 \\
			
			\bottomrule
	\end{tabular}}
\end{table*}

\subsection{Evaluation Metrics}
We adopt four evaluation metrics to evaluate the performance of various methods, including dice similarity coefficient (DSC), Jaccard index (JAC), Hausdorff distance (HSD) and mean absolute error (MAE). Dice similarity coefficient is a statistic used to gauge the similarity between two sets of data. Jaccard index is a measure used to evaluate the performance of an algorithm in the field of detecting objects within an image. Hausdorff distance is the maximum distance of a set to the nearest point in the other set. Mean absolute error is a quantitative measure of the average difference between all elements in the predicted mask and the ground truth. The higher value is better for dice similarity coefficient and Jaccard index while the lower is better for Hausdorff distance and mean absolute error.

\subsection{Implementation Details}
We implement our proposed method on PyTorch framework and conduct all the experiments on a single NVIDIA RTX 4090 GPU with fixed random seeds. For CNN-based students, we choose the semantic FPN \cite{kirillov2019panoptic} as decoder with encoders of MobileNetV2 \cite{sandler2018mobilenetv2} and ResNet-50 \cite{he2016deep}; for Transformer-based students, we utilize the efficient SegFormer \cite{xie2021segformer} decoder with encoders of MiT-B0 and MiT-B2, which have comparable parameters with their CNN counterparts, respectively. Pre-trained weights on ImageNet-1k \cite{deng2009imagenet} are employed to initialize the backbone networks. We itemize the crop size (cs), batch size (bs), learning rate (lr), training epochs (ep), optimizer (opt) for each dataset: (1) Synapse Multi-organ dataset and ACDC dataset: cs=224$\times$224; bs=8; lr=0.0003; ep=200; opt=AdamW. (2) Kvasir dataset: cs=352$\times$352; bs=8; lr=0.00005; ep=300; opt=AdamW. We utilize the polynomial learning-rate strategy \cite{liu2015parsenet} to adjust the learning rate with the power of 0.9. We adopt the random flip and random rotation for data augmentation to avoid overfitting. For a more robust and fair comparison, we take the checkpoint from the last epoch and report the results on the testing sets. 

\begin{figure}[t]
	\centering 
	\includegraphics [width=\linewidth] {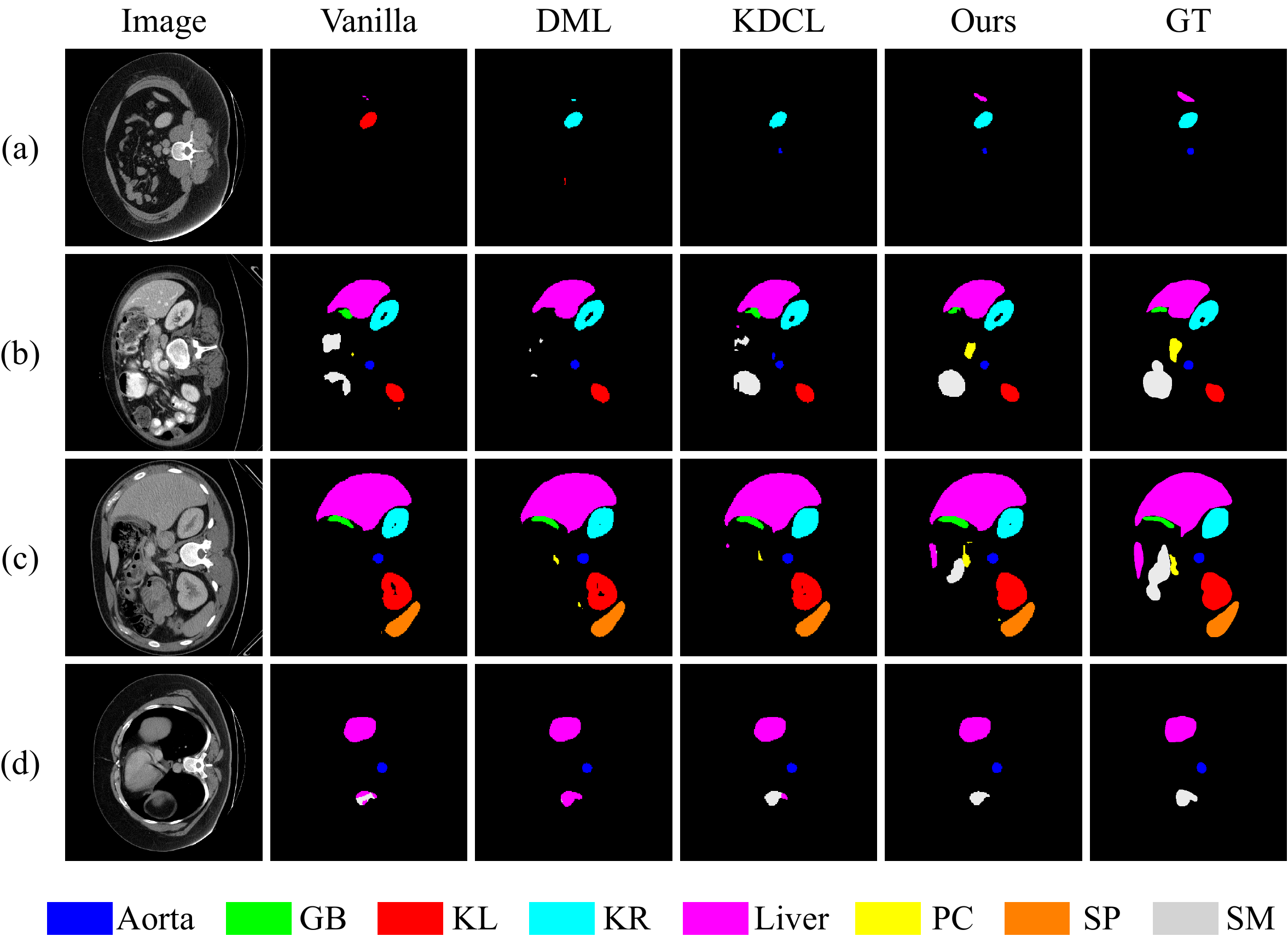}
	\caption{Visual comparisons on Synapse Multi-organ dataset. (a) MobileNetV2. (b) MiT-B0. (c) ResNet-50. (d) MiT-B2.}
	\label{fig:visual_synapse}
\end{figure}  

\begin{figure}[t]
	\centering 
	\includegraphics [width=\linewidth] {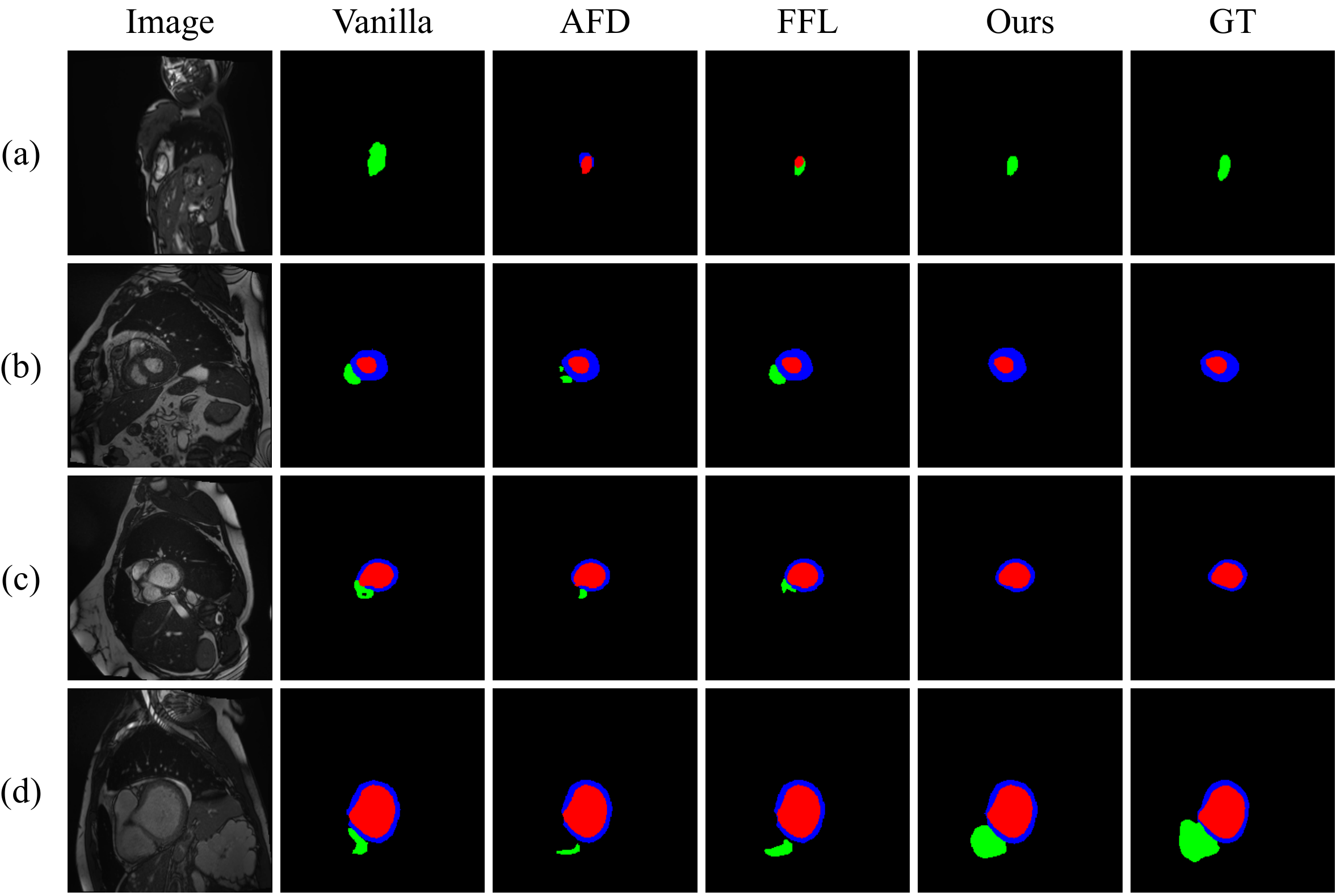}
	\caption{Visual comparisons on ACDC dataset. Green, red and blue denote RV, LV and Myo respectively. (a) MobileNetV2. (b) MiT-B0. (c) ResNet-50. (d) MiT-B2.}
	\label{fig:visual_ACDC}
\end{figure}

\begin{figure}[t]
	\centering 
	\includegraphics [width=\linewidth] {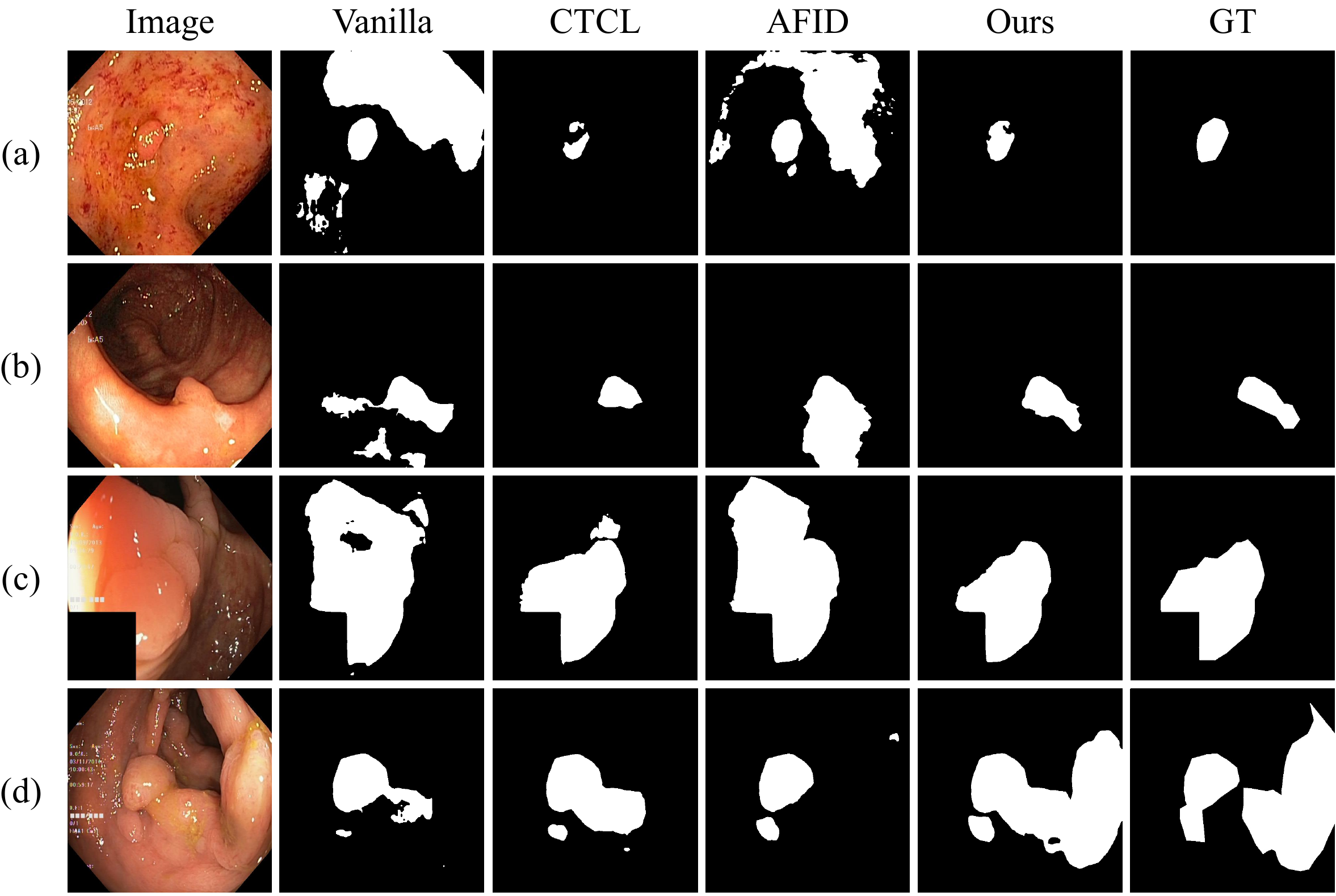}
	\caption{Visual comparisons on Kvasir dataset. The polyps are shown in white. (a) MobileNetV2. (b) MiT-B0. (c) ResNet-50. (d) MiT-B2.}
	\label{fig:visual_polyp}
\end{figure}

\subsection{Comparisons with State-of-the-arts}
We compare our proposed method with other 6 state-of-the-art collaborative learning methods, including DML \cite{zhang2018deep}, AFD \cite{chung2020feature}, CTCL \cite{zhu2023good}, KDCL \cite{guo2020online}, FFL \cite{kim2021feature}, and AFID \cite{su2021attention}. Among them, DML, AFD and CTCL are mutual learning based methods while KDCL, FFL and AFID are ensemble learning based methods. For comparison, we implement all the methods on two pairs: MobileNetV2 and MiT-B0; ResNet-50 and MiT-B2. Experimental results and analyses are as follows.

\subsubsection{Results on Synapse Multi-organ Dataset}
We first evaluate the aforementioned methods on the synapse multi-organ dataset and report the quantitative results in Table \ref{table:synapse and ACDC}. In contrast to the students without KD (Vanilla), all the methods show varying improvements in terms of DSC, JAC and HSD while our method achieves consistently the best performance compared to other online KD methods. Specifically, our method produces a remarkable margin of 3.74\%, 4.91\%, 31.33\% on MobileNetV2 and 3.11\%, 4.05\%, 18.51\% on MiT-B0 in DSC, JAC and HSD, respectively, much outperforming the second best method CTCL which improves MobileNetV2 by 2.26\%, 3.11\%, 12.91\% and MiT-B0 by 1.74\%, 2.30\%, 11.14\% in terms of DSC, JAC and HSD. Moreover, our method achieves larger benefits for more sophisticated students, with an improvement of 3.99\%, 5.83\%, 27.33\% on ResNet-50 and 3.80\%, 6.31\%, 25.68\% on MiT-B2 in DSC, JAC and HSD, respectively, indicating that reasonable knowledge transferring contributes to stimulating the potential of complicated student models especially under the condition of limited training data. 

\subsubsection{Results on ACDC Dataset}
To avoid randomness, we also conduct our experiments on the ACDC dataset and the quantitative results are shown in Table \ref{table:synapse and ACDC}. As expected, our method still achieves the best segmentation performance over other online KD methods. Specifically, our methods outperforms the previous best mutual learning based method CTCL with improvements of 0.91\%, 1.51\%, 10.22\% on MobileNetV2 and 0.97\%, 1.61\%, 4.10\% on MiT-B0 in terms of DSC, JAC and HSD, respectively. In addition, compared with the best ensemble learning based method AFID, our method improves the evaluation of DSC, JAC by 0.85\%, 1.59\% on ResNet-50 and 1.34\%, 2.36\% on MiT-B2.  

\subsubsection{Results on Kvasir Dataset}
To verify the generalization ability of our proposed method, we conduct the experiments on Kvasir dataset and report the quantitative results in Table \ref{table:polyp}. Different from the synapse multi-organ and ACDC datasets, there is only one class to be segmented in the Kvasir dataset, leading to limited category information. Besides, compared to the organ segmentation tasks with strong position and shape priors, there is a large variation of position and shape for polyps in endoscopic images, increasing the difficulty of accurate polyp segmentation. Surprisingly, nearly all the online KD methods achieve more prominent improvements on both CNN and Transformer students in contrast to their performances on synapse multi-organ and ACDC datasets due to the complementary intrinsic properties of Transformer and CNN. Meanwhile, our method still achieves the highest performance with improvements of 1.22\%, 1.84\%, 11.56\% on MobileNetV2 and 1.13\%, 1.44\%, 5.3\% on MiT-B0; 1.67\%, 1.99\%, 8.4\% on ResNet-50 and 1.36\%, 1.46\%, 16.95\% on MiT-B2 over the previous best model CTCL in terms of DSC, JAC and MAE, respectively. These quantitative results on three datasets substantiate the fine robustness of our proposed RCL.
\subsubsection{Visual Comparison}
We also show some visual comparisons of our proposed method and other state-of-the-art online KD methods on Synapse Multi-organ dataset, ACDC dataset and Kvasir dataset in Fig. \ref{fig:visual_synapse}, \ref{fig:visual_ACDC} and \ref{fig:visual_polyp} respectively to exhibit the superiority of our approach from an intuitive view. As we can see, compared to the vanilla students (Column 2), all the online KD methods (Column 3-5) achieve qualitative improvements on the segmentation maps. Meanwhile, our method (Column 5) is capable of generating fine predictions that are more consistent with the ground truth (Column 6) in contrast to either mutual learning based method (Column 3) or ensemble learning based method (Column 4). Specifically, with our method implemented, there are more accurate locations of the segmented objects for CNN-based students MobileNetV2 (Row 1) and ResNet-50 (Row 3), and more detailed boundary descriptions of them for Transformer-based students MiT-B0 (Row 2) and MiT-B2 (Row 4), which indicates the effectiveness of the bidirectional knowledge transmission between CNN student and Transformer student achieved by our proposed framework.

\subsection{Ablation Studies}
The proposed framework mainly consists of two key modules, namely rectified logit-wise collaborative learning (RLCL) and class-aware feature-wise collaborative learning (CFCL). In this section, we carry out a series of experiments with different settings to verify the performance of different components and key modules. We conduct our ablation studies on the Synapse Multi-organ dataset using ResNet-50 and MiT-B2, and choose DSC and HSD for evaluation.
\subsubsection{The Effectiveness of Different Components}
Table \ref{table:ablition_all} reports the results of different components. As we can see, both RLCL and CFCL can enhance the performance of CNN-based and Transformer-based students. Specifically, it represents an improvement of 2.76\%, 22.45\% on ResNet-50 and 2.53\%, 19.08\% on MiT-B2 in terms of DSC and HSD as we employ the RLCL on the vanilla students. Similarly, with the CFCL implemented, there is a gain of 1.73\%, 14.51\% on ResNet-50 and 1.76\%, 13.05\% on MiT-B2 in DSC and HSD, respectively. The best results are achieved by combining both RLCL and CFCL, which demonstrates a good cooperativity of the proposed two modules.
\begin{table}[t]	
	\caption{Ablation Analysis of Different Components. And the Best Results are Shown in \textbf{Boldface}.}
	\centering
	\label{table:ablition_all}
	\scalebox{1}{
		\begin{tabular}{cccccccc}
			\toprule
			\multirow{2}{*}{No.}  &\multirow{2}{*}{Vanilla}  &\multirow{2}{*}{RLCL}   &\multirow{2}{*}{CFCL}   & \multicolumn{2}{c}{ResNet-50}	&\multicolumn{2}{c}{MiT-B2} 	  \\
			\cmidrule(r){5-6}       \cmidrule(r){7-8}
			{~}                  &{~}              &{~}                 &{~}               &DSC$\uparrow$    &HSD$\downarrow$              &DSC$\uparrow$    &HSD$\downarrow$      \\       
			\midrule
			1                   &\checkmark       &                    &                  &75.62    &30.11             &77.85    &26.52                       \\		
			2                   &\checkmark		&\checkmark	         &                  &77.71   &23.35             &79.82    &21.46        \\
			3                   &\checkmark       &                   &\checkmark        &76.93    &25.74             &79.22    &23.06                    \\
			4                   &\checkmark       &\checkmark         &\checkmark        &\textbf{78.64}    &\textbf{21.88}          &\textbf{80.81}    &\textbf{19.71}            \\
			
			\bottomrule
	\end{tabular}}
\end{table}
\begin{figure}[t]
	\centering 
	\includegraphics [width=\linewidth] {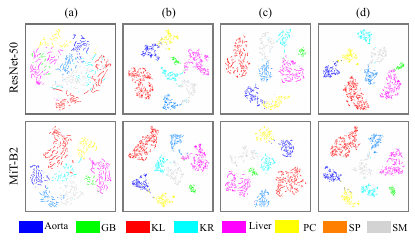}
	\caption{t-SNE visualization of the extracted features from CNN-based and Transformer-based students with our proposed RLCL and CFCL. (a) Vanilla. (b) Vanilla + RLCL. (c) Vanilla + CFCL. (d) Vanilla + RLCL + CFCL.}
	\label{fig:overa_tsne}
\end{figure}

In Fig. \ref{fig:overa_tsne}, we visualize the feature representations extracted from two students with different components. As we can see, our proposed RLCL and CFCL encourage significant inter-class separation and intra-class compactness, indicating that our proposed methods effectively help the student models to learn discriminative features for each class, which is crucial for segmentation predictions.
\begin{table}[t]	
	\caption{Comparison of Our RLCL with Other Logit-wise Collaborative Learning Strategies. The Best Results are Shown in \textbf{Boldface}.}
	\centering
	\label{table:ablition_RLCL}
	\scalebox{1}{
		\begin{tabular}{clcccc}
			\toprule
			\multirow{2}{*}{No.}	&\multirow{2}{*}{\quad\quad\, Methods}      & \multicolumn{2}{c}{ResNet-50}	&\multicolumn{2}{c}{MiT-B2} 	  \\
			\cmidrule(r){3-4}       \cmidrule(r){5-6}
			{~}                            &{~}                      &DSC$\uparrow$    &HSD$\downarrow$              &DSC$\uparrow$    &HSD$\downarrow$    \\       
			\midrule
			1		 &\quad Vanilla                               &75.62    &30.11             &77.85    &26.52                       \\		
			2		 &\quad+ ML\cite{zhang2018deep}                                            &76.33    &27.45             &78.09   &25.13                   \\
			3		 &\quad+ EL \cite{guo2020online}                                            &76.55    &26.87             &78.25    &24.63                    \\
			4		 &\quad+ BSD \cite{zhu2023good}                                             &76.89    &25.62             &78.93    &23.15                   \\
			5		 &\quad+ RLCL (Ours)\quad\quad                               &\textbf{77.71}    &\textbf{23.35}          &\textbf{79.82}    &\textbf{21.46}            \\
			
			\bottomrule
	\end{tabular}}
\end{table}
\begin{table}[t]	
	\caption{Ablation Analysis of the Rectification Weight $\lambda$. The Best Results are Shown in \textbf{Boldface}.}
	%\vspace{0.15cm}
	\centering
	\label{table:ablition_lambda}
	\scalebox{1}{
		\begin{tabular}{clcccc}
			\toprule
			\multirow{2}{*}{No.}	&\multirow{2}{*}{\quad\quad\quad\quad$\lambda$}      & \multicolumn{2}{c}{ResNet-50}	&\multicolumn{2}{c}{MiT-B2} 	  \\
			\cmidrule(r){3-4}       \cmidrule(r){5-6}
			{~}                            &{~}                      &DSC$\uparrow$    &HSD$\downarrow$              &DSC$\uparrow$    &HSD$\downarrow$    \\       
			\midrule
			1		 &\quad$\lambda = 1$                               &75.62    &30.11             &77.85    &26.52                       \\		
			2		 &\quad$\lambda = \lambda^a$                                 &76.71    &26.24             &78.82   &23.67                   \\
			3		 &\quad$\lambda = \lambda^a \cdot \lambda^s$                 &77.42    &24.68             &79.57    &21.94                   \\
			4		 &\quad$\lambda = \lambda^a \cdot \lambda^c$                 &77.14    &25.16             &79.28    &22.65                   \\
			5		 &\quad$\lambda = \lambda^a \cdot \lambda^s \cdot \lambda^c$\quad\quad   &\textbf{77.71}    &\textbf{23.35}          &\textbf{79.82}    &\textbf{21.46}            \\
			
			\bottomrule
	\end{tabular}}
\end{table}
\begin{table}[!t]
	\caption{Ablation Analysis of CFCL on Different Stages. ${\mathcal{L}}_{cfcl}^{E}$ and ${\mathcal{L}}_{cfcl}^{D}$ Denote the Class-aware feature-wise distillation on the Output Features of Encoder and Decoder, Respectively. The Best Results are Shown in \textbf{Boldface}.}
	\centering
	\label{table:ablition_CFCL}
	\scalebox{1}{
		\begin{tabular}{cccccccc}
			\toprule
			\multirow{2}{*}{No.} &\multirow{2}{*}{Vanilla}  &\multirow{2}{*}{${\mathcal{L}}_{cfcl}^{E}$}   &\multirow{2}{*}{${\mathcal{L}}_{cfcl}^{D}$}   & \multicolumn{2}{c}{ResNet-50}	&\multicolumn{2}{c}{MiT-B2} 	  \\
			\cmidrule(r){5-6}       \cmidrule(r){7-8}
			{~}                &{~}                 &{~}                 &{~}               &DSC$\uparrow$    &HSD$\downarrow$              &DSC$\uparrow$    &HSD$\downarrow$      \\       
			\midrule
			1               &\checkmark           &                    &                  &75.62    &30.11             &77.85    &26.52                       \\		
			2               &\checkmark    	&\checkmark	         &                  &76.54   &26.67             &78.24    &25.14        \\
			3                &\checkmark          &                   &\checkmark        &76.18    &28.32             &78.71    &24.21                    \\
			4               &\checkmark          &\checkmark         &\checkmark        &\textbf{76.93}    &\textbf{25.74}          &\textbf{79.22}    &\textbf{23.06}            \\
			
			\bottomrule
	\end{tabular}}
\end{table}
\subsubsection{The Effectiveness of RLCL}
In Table \ref{table:ablition_RLCL}, we compare the RLCL with three logit-wise collaborative learning strategies including mutual learning (ML) \cite{zhang2018deep}, ensemble learning (EL) \cite{guo2020online} and bi-directional selective distillation (BSD) \cite{zhu2023good}. As we can see, ML achieves inferior improvements on both students for overlooking the accuracy of student soft labels. EL obtains a considerable increase due to the enhanced ensemble label. BSD selects the reliable regions in student soft labels for bidirectional distillation and attains a large improvement. Comparatively, our RLCL achieves the best performance on both two students, indicating its superiority that adaptively rectifies the wrong regions in student soft labels for accurate knowledge transfer in the logit space.

We also explore the influence of the rectification weight $\lambda$ with different configurations and report the quantitative results in Table \ref{table:ablition_lambda}. Compared to the intuitive logit-wise collaborative learning without rectification ($\lambda$ = 1), an obvious improvement is achieved as we align the probabilities of mis-categorized class and truth class for the wrong regions of soft labels via $\lambda^a$. In addition, we observe that both similarity-based decay factor $\lambda^s$ and certainty-based decay factor $\lambda^c$ can further enhance the performance of two students due to the adaptive rectification under the guidance of ground truth. By combining $\lambda^a$, $\lambda^s$ and $\lambda^c$ as the final rectification weight $\lambda$, we obtain the best results on both two students through logit-wise collaborative learning with dynamically rectified soft labels.  

\subsubsection{The Effectiveness of CFCL}
Table \ref{table:ablition_CFCL} demonstrates the effectiveness of CFCL on the output features of encoder and decoder. As we can see, both the CFCL on the output feature of encoder and that of decoder contribute to a performance improvement on two students. Meanwhile, we observe that the CFCL on the output feature of encoder benefits more than that of the decoder for CNN-based student (DSC: +1.22\% vs. +0.74\%) while this circumstance is on the contrary for Transformer-based student (DSC: +0.5\% vs. +1.10\%). That is because the output feature of Transformer encoder incorporates more global semantic information that benefits the CNN-based student for accurate location while the output feature of CNN decoder is richer in local detail information, encouraging the Transformer-based student to predict more elaborate boundaries.

\subsubsection{The Sensitivity of $\beta$, $\gamma_{1}$ and $\gamma_{2}$}
Table \ref{table:ablition_hyperparameters} reports the DSC(\%) of the students with different ratios of $\beta$, $\gamma_{1}$ and $\gamma_{2}$ on the Synapse Multi-organ dataset. As we can see, increasing the importance of $\beta$ significantly improves the performance of CNN-based student due to the accurate knowledge transfer from Transformer to CNN by the proposed RLCL. However, the performance of Transformer-based student degrades slightly when $\beta$=5. As our goal is to facilitate the collaborative learning between two students, we choose $\beta$=3 as it presents the best trade-off for both students. Similarly, we choose $\gamma_{1}$=1 and $\gamma_{2}$=2 for the proposed CFCL between CNN-based student and Transformer-based student on the output features of encoder and decoder, respectively. For the sake of convenience and robustness, we adopt the configuration of $\beta$=3, $\gamma_{1}$=1 and $\gamma_{2}$=2 for all the experiments of our method, i.e., two pairs of CNN-Transformer students on three medical image segmentation datasets.

\begin{table}[t]

	\caption{Ablation Analysis of $\beta$, $\gamma_{1}$ and $\gamma_{2}$. $\Delta_{s}$ is the Sum of Improvement in DSC on Both Students Compared with the Vanilla.}
	\centering
	\label{table:ablition_hyperparameters}
	\scalebox{1}{
		\begin{tabular}{ccccccc}
			\toprule
			$\beta$           &0                 &1                  &2                      &3                      &4                                 &5 \\
			\midrule
			ResNet-50                &75.62    &77.15            &77.56                   &77.71    &77.90             &77.34                           \\		
			MiT-B2               &77.85    	&78.84	         &79.09                  &79.82   &78.57             &77.64            \\
			$\Delta_{s}$                &0          &+2.52         &+3.18        &\textbf{+4.06}    &+3.00             &+1.51                        \\
			\midrule
			$\gamma_{1}$           &0                 &0.1                  &0.5                      &1                      &2                                 &5 \\
			\midrule
			ResNet-50                &75.62    &75.87            &76.17                   &76.54    &76.63             &76.04                           \\		
			MiT-B2               &77.85    	&77.96	         &78.38                  &78.24   &77.94             &77.52            \\
			$\Delta_{s}$                &0          &+0.36         &+1.08        &\textbf{+1.31}    &+1.10             &+0.09                        \\
			\midrule
			$\gamma_{2}$           &0                 &1                  &2                      &3                      &4                                 &5 \\
			\midrule
			ResNet-50                &75.62    &75.95            &76.18           &76.21    &75.85             &75.47                           \\		
			MiT-B2               &77.85    	&78.56	         &78.71                  &78.34   &78.33            &78.22            \\
			$\Delta_{s}$                &0          &+1.04         &\textbf{+1.42}        & +1.08   &+0.71             &+0.32                        \\
			
			\bottomrule
	\end{tabular}}
\end{table}

\subsection{Generalization Studies} 
In this section, we design a series of extensive experiments to demonstrate the generalization of our method. The results and analyses are as follows. 
\subsubsection{Generalization on Different Pairs of Students}
\begin{figure}[!t]
	\centering 
	\includegraphics [width=\linewidth] {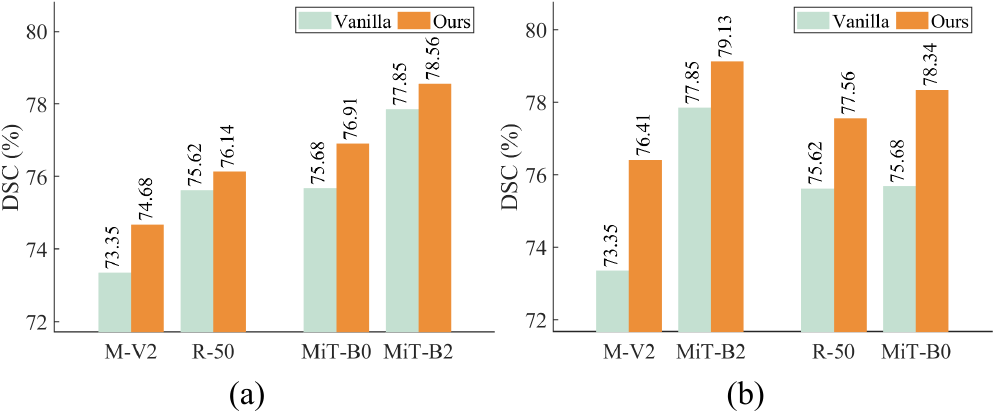}
	\caption{Results of our method on different pairs of students. (a) Isomorphic students (MobileNetV2 \& ResNet-50; MiT-B0 \& MiT-B2). (b) Heterogeneous students in different sizes (MobileNetV2 \& MiT-B2; ResNet-50 \& MiT-B0).}
	\label{fig:different_match}
	%\vspace{-0.3cm}
\end{figure}

Fig.~\ref{fig:different_match} presents the results of our method on different pairs of students. Specifically, we implement our method on isomorphic students in Fig.~\ref{fig:different_match}~(a) and heterogeneous students with incomparable parameters in Fig. \ref{fig:different_match}~(b). As we can see, our method still achieves considerable improvements in DSC for sub-students, indicating the outstanding generalization and robustness of our method. Moreover, CNN-Transformer collaborative learning benefits more than CNN-CNN or Transformer-Transformer collaborative learning due to the global-local complementarity between CNN and Transformer. 
\subsubsection{Generalization on Knowledge Distillation} 
\begin{figure}[!t]
	\centering 
	\includegraphics [width=0.82\linewidth] {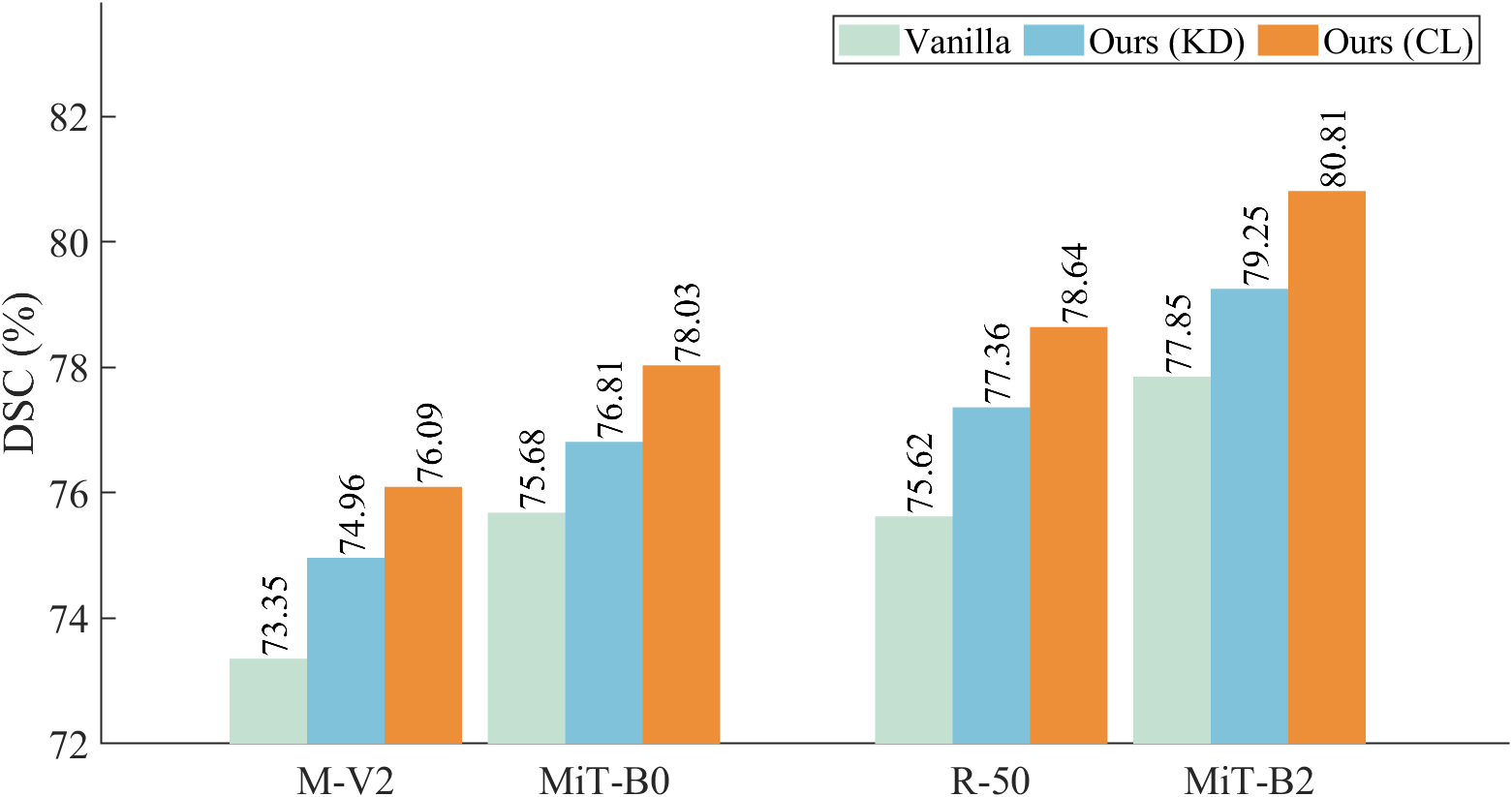}
	\caption{Results of our method on knowledge distillation (KD) and comparison with collaborative learning (CL).}
	\label{fig:ctrkd}
	%\vspace{-0.3cm}
\end{figure}

We also migrate our method to knowledge distillation (KD), i.e., the CNN-based and Transformer-based students guiding each other as teachers respectively in an offline manner. The comparison results with collaborative learning (CL) are shown in Fig. \ref{fig:ctrkd}. Obviously, our method can yet enhance the performance of students in offline KD paradigm, further verifying the generalization ability of our method. Meanwhile, we observe that simultaneous collaborative learning achieves larger improvements than two-stage KD for both CNN-based and Transformer-based students because the teacher is progressively optimized in the collaborative learning process. 
\begin{figure}[!t]
	\centering 
	\includegraphics [width=0.95\linewidth] {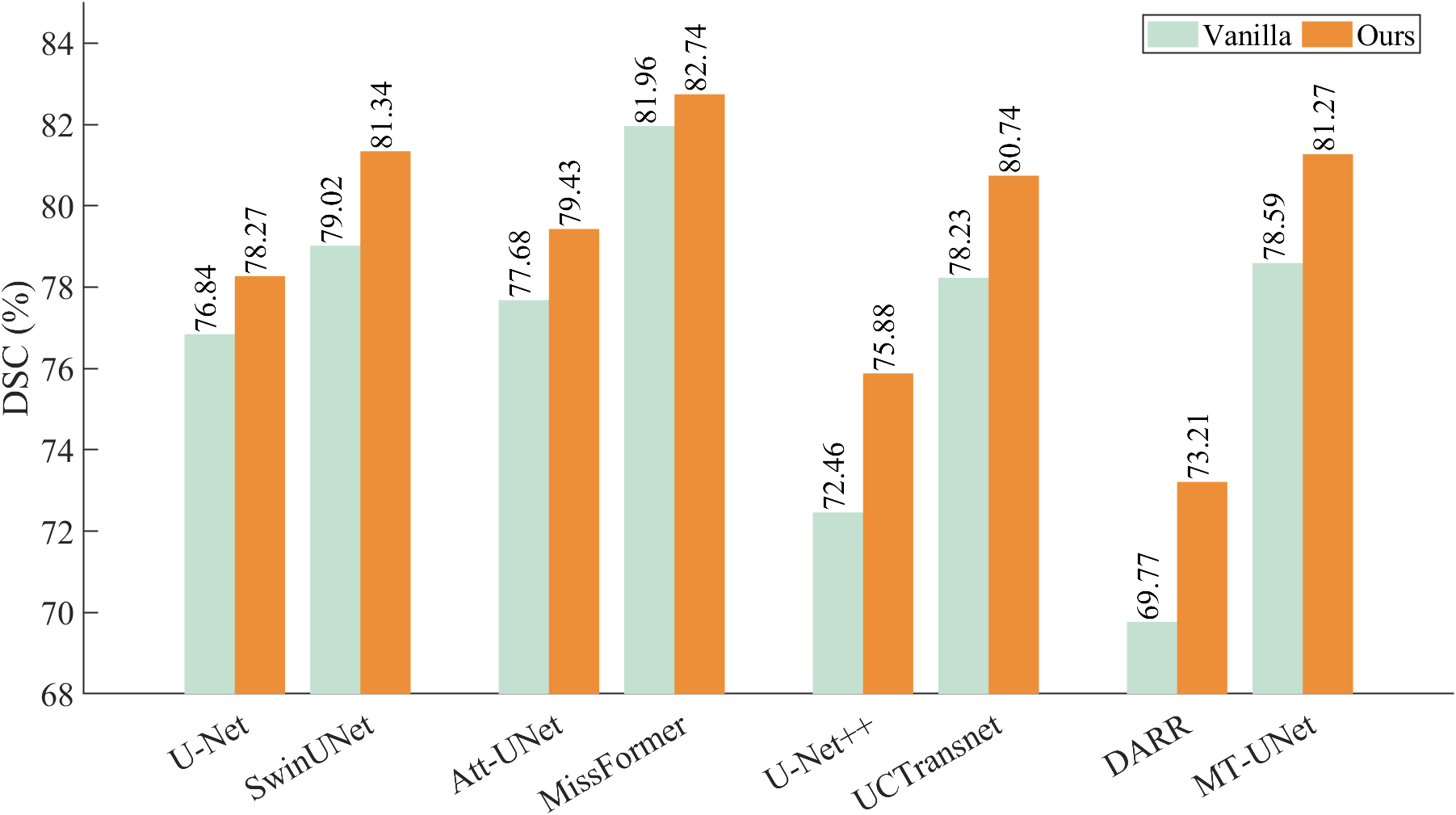}
	\caption{Results of our method on existing CNN-based and Transformer-based MIS methods.}
	\label{fig:othermethod}
	%\vspace{-0.3cm}
\end{figure}
\subsubsection{Generalization on Existing MIS Methods}
In Fig.~\ref{fig:othermethod}, we extend our proposed framework to existing CNN-based and Transformer-based MIS methods, including 4 pairs of CNN-Transformer students: U-Net \cite{ronneberger2015u} \& SwinUNet \cite{cao2022swin}; Att-UNet \cite{schlemper2019attention} \& MissFormer \cite{huang2022missformer}; U-Net++ \cite{zhou2018unet++} \& UCTransnet \cite{wang2022uctransnet}; DARR \cite{fu2020domain} \& MT-UNet \cite{wang2022mixed}. Our method presents excellent generalization ability on various students and achieves consistent improvements on them. The results indicate that our framework can significantly improve the performance of existing networks via accurate and effective collaborative learning between them, which makes great sense to the development of MIS models in the future.

\section{Conclusion}
In this paper, we propose a CNN-Transformer rectified collaborative learning (CTRCL) framework to learn stronger CNN-based and Transformer-based models for MIS tasks via bi-directional knowledge transfer between them in both logit and feature spaces. For accurate logit-wise knowledge transfer, we propose an RLCL strategy which introduces the ground truth to adaptively rectify the wrong regions in soft labels. For effective feature-wise knowledge transfer between heterogeneous CNN and Transformer, we propose a CFCL strategy to grant their immediate features the similar category perception ability. Experimental results on three popular benchmarks show superior performance of our method over 6 state-of-the-art collaborative learning methods. Furthermore, extensive ablation and generalization studies demonstrate the effectiveness of each component as well as the generalization ability of our method.

%\section*{Acknowledgments}
%This should be a simple paragraph before the References to thank those individuals and institutions who have supported your work on this article.
%
%

 
% {\small
% 	% \bibliographystyle{ieee_fullname}
% 	\bibliographystyle{ieeetran}
% 	\bibliography{ref_fsl}
% }

%\newpage

% your own custom macro containing the $\backslash${\tt{includegraphics}} command to make things
%\bf{If you include a photo:}\vspace{-33pt}
\begin{IEEEbiography}[{\includegraphics[width=25mm,height=35mm,clip,keepaspectratio]{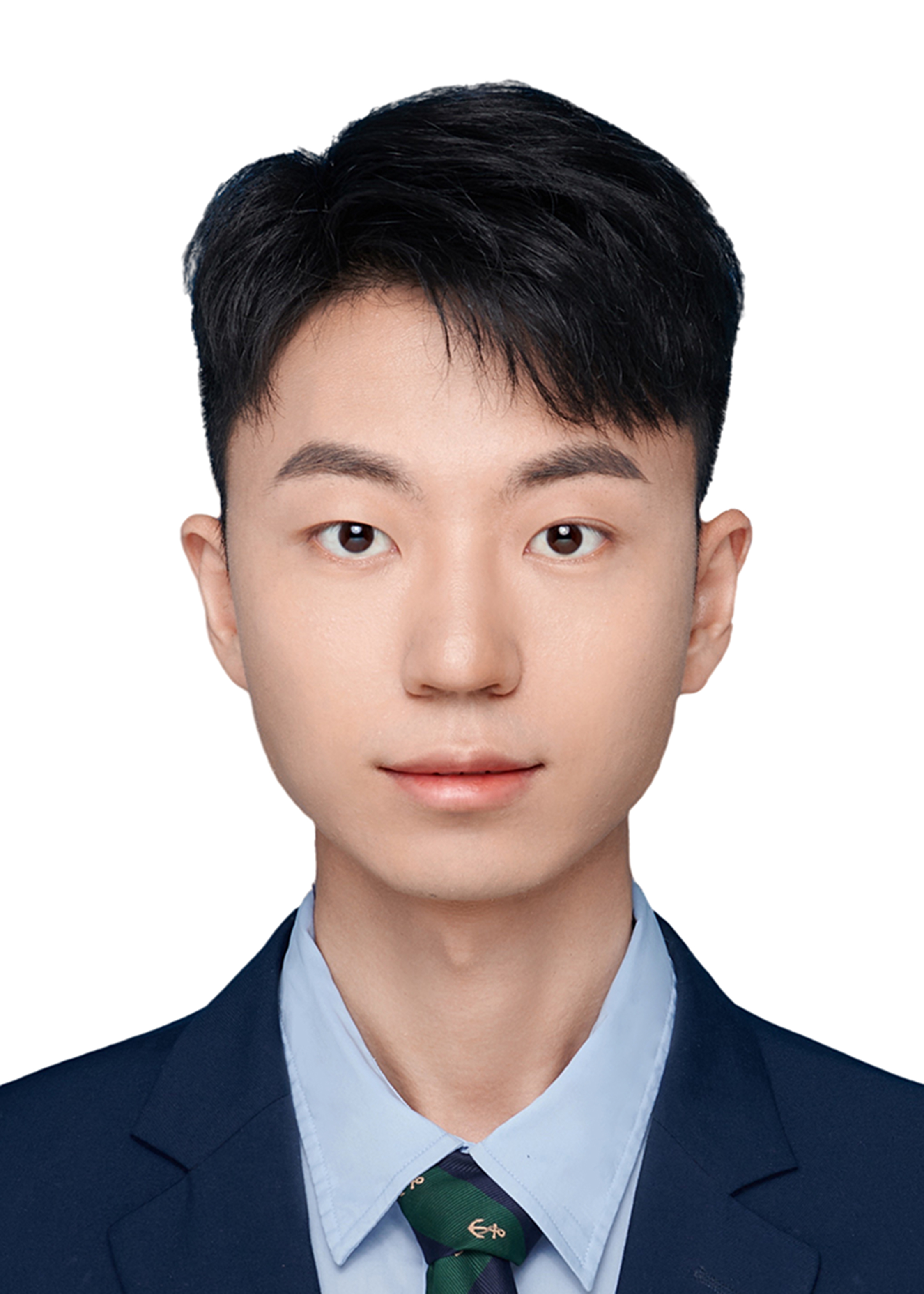}}]{Lanhu Wu}
	received the B.S. degree in communication engineering from Dalian University of Technology, Dalian, China, in 2020. 

    He is currently with IIAU-OIP Lab, Dalian University of Technology, Dalian, China. His current research interests include computer vision, medical image segmentation and model compression.
\end{IEEEbiography}
\vspace{-1cm}
\begin{IEEEbiography}[{\includegraphics[width=25mm,height=35mm,clip,keepaspectratio]{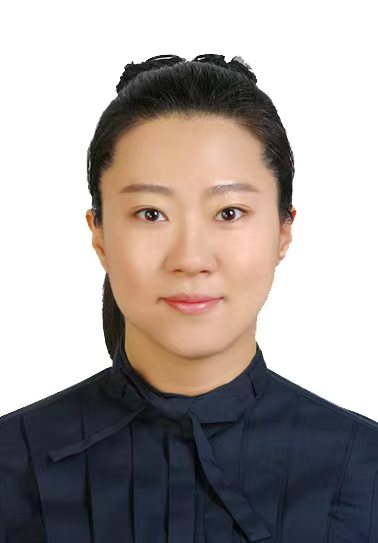}}]{Miao Zhang}
	(Member, IEEE) received the B.S. degree in computer science from the Memorial University of Newfoundland, St. John’s, NL, Canada, in 2005, and the Ph.D. degree in electronic engineering from Kwangwoon University, Seoul, South Korea, in 2012. From 2013 to 2015, she was an Assistant Professor with the Department of Game and Mobile Contents, Keimyung University, Daegu, South Korea, and an Adjunct Professor with the Department of Computer Science, DigiPen Institute of Technology, Redmond, WA, USA, respectively.
	
	She is currently an Associate Professor with the Key Laboratory for Ubiquitous Network and Service Software of Liaoning Province, DUT-RU International School of Information Science and Software Engineering, Dalian University of Technology, Dalian, China. Her research interests include computer vision, machine learning, and 3D imaging and visualization.
\end{IEEEbiography}
\vspace{-1cm}
\begin{IEEEbiography}[{\includegraphics[width=25mm,height=35mm,clip,keepaspectratio]{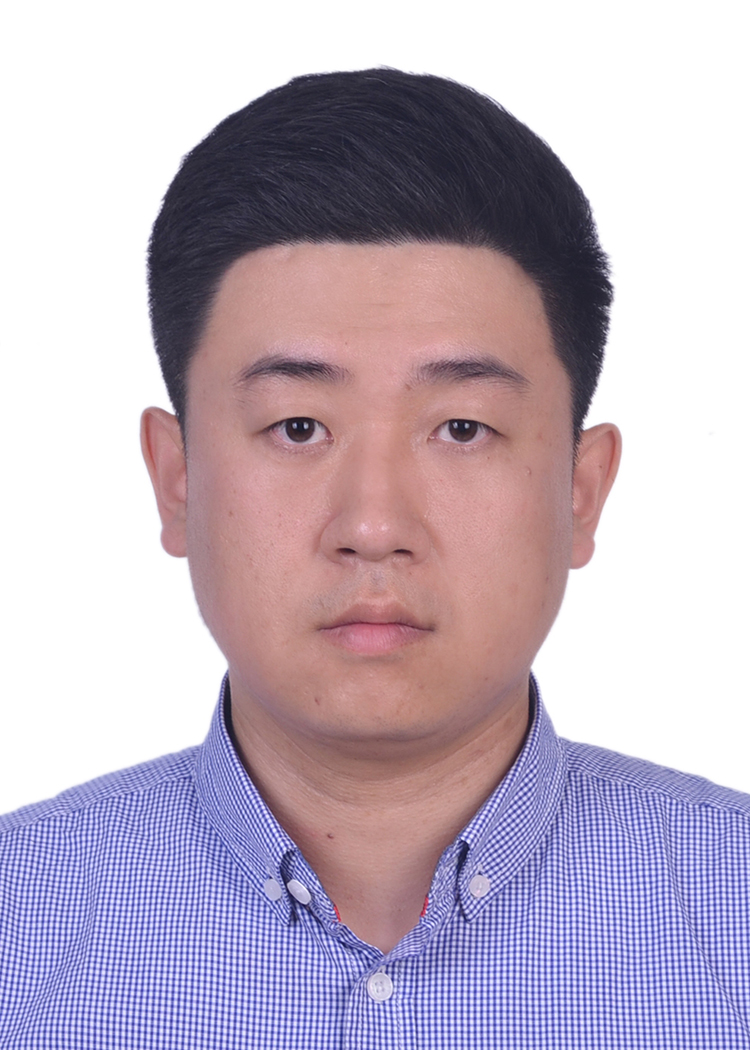}}]{Yongri Piao}
	((Member, IEEE) received the M.Sc. and Ph.D. degrees in information and communication engineering from Pukyong National University, Busan, South Korea, in 2005 and 2008, respectively. 
	
	Since 2012, he has been an Associate Professor of information and communication engineering with the Dalian University of Technology, Dalian, China. His research interests include 3D computer vision and sensing, object detection and target recognition, and 3D reconstruction and visualization.
\end{IEEEbiography}
\vspace{-0.5cm}
\begin{IEEEbiography}[{\includegraphics[width=25mm,height=35mm,clip,keepaspectratio]{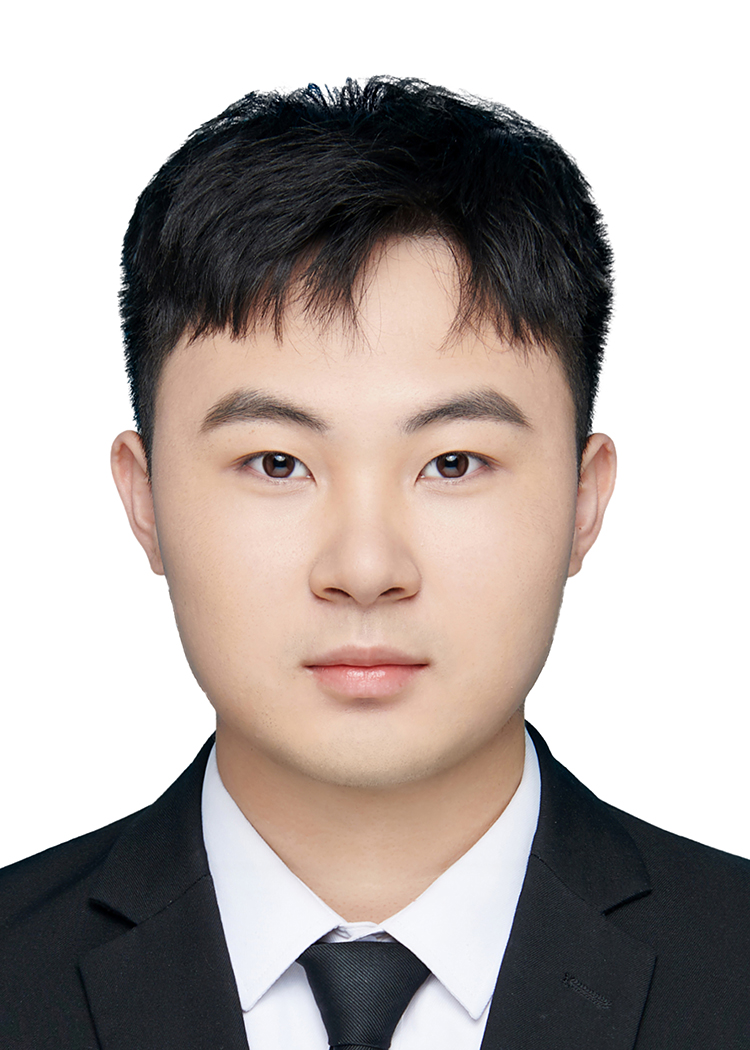}}]{Zhenyan Yao}
	received the B.S. degree in communication engineering from North China Electric Power University, Baoding, China, in 2022. 
	
	He is currently with IIAU-OIP Lab, Dalian University of Technology, Dalian, China. His current research interests include computer vision, medical image segmentation and semi-supervised learning.
\end{IEEEbiography}
\vspace{-0.2cm}
\begin{IEEEbiography}[{\includegraphics[width=25mm,height=35mm,clip,keepaspectratio]{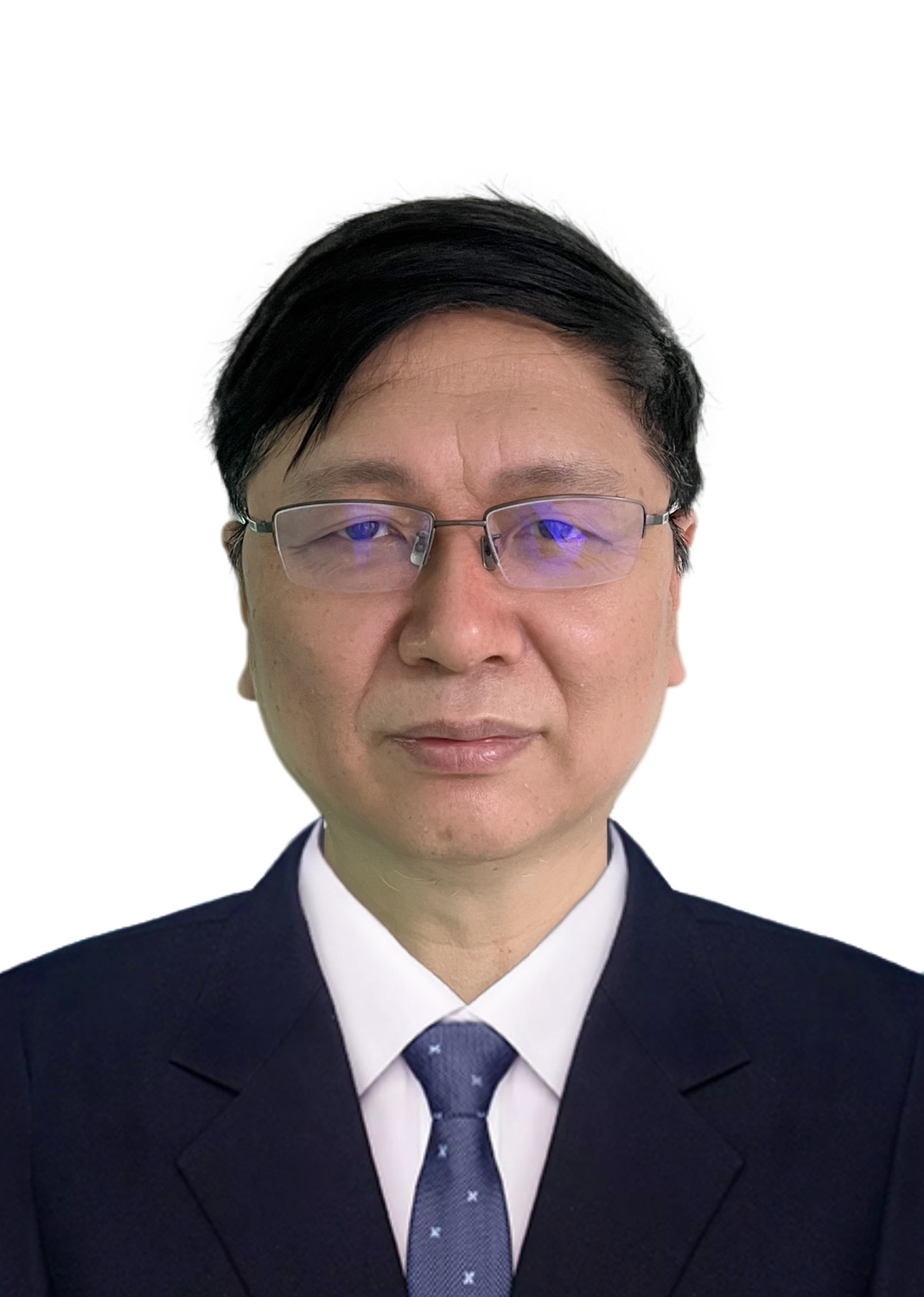}}]{Weibing Sun}
	Affiliated Zhongshan Hospital of Dalian University, Urology, Chief physician, Professor.
\end{IEEEbiography}

\begin{IEEEbiography}[{\includegraphics[width=25mm,height=35mm,clip,keepaspectratio]{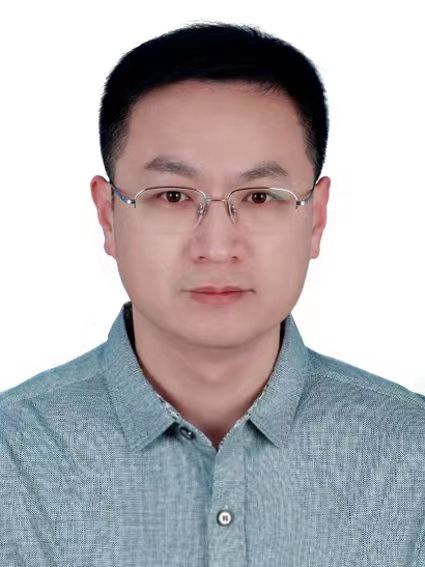}}]{Feng Tian}
	Affiliated Zhongshan Hospital of Dalian University, Urology, Chief physician, Professor.
\end{IEEEbiography}

\begin{IEEEbiography}[{\includegraphics[width=25mm,height=35mm,clip,keepaspectratio]{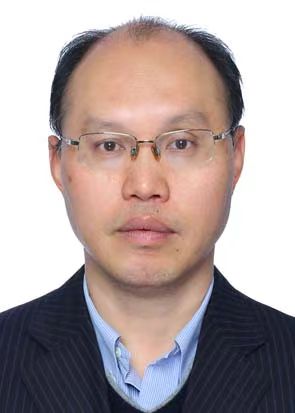}}]{Huchuan Lu}
	(Fellow, IEEE) received the M.S. degree in signal and information processing and the Ph.D. degree in system engineering from the Dalian University of Technology (DUT), Dalian, China, in 1998 and 2008, respectively. 
	
	He joined the Faculty of the School of Information and Communication Engineering, DUT, in 1998, where he is currently a Full Professor. His research interests include computer vision and pattern recognition with a focus on visual tracking, saliency detection, and segmentation.
\end{IEEEbiography}


\vfill

\end{document}